\documentclass[11pt,preprint]{imsart}

\usepackage{amsmath}
\usepackage{amssymb}
\usepackage{amsfonts}
\usepackage{anysize}
\usepackage{epsfig}
\usepackage{natbib}
\usepackage{subfigure}
\usepackage{multirow}
\usepackage{graphicx}
\usepackage{algorithm}
\usepackage{algorithmic}

\RequirePackage[OT1]{fontenc}
\RequirePackage{amsthm,amsmath}
\RequirePackage[colorlinks,citecolor=blue,urlcolor=blue]{hyperref}
\RequirePackage{hypernat}

\newcommand{\realline}{\mathbb{R}}
\newcommand{\mysec}[1]{Section~\ref{sec:#1}}

\newcommand{\E}{\mathbb{E}}

\newcommand{\DP}{\textrm{DP}}

\def\e{\mathrm{e}}
\def\bK{\boldsymbol{\mathrm{K}}}

\def\bX{{\mathrm{X}}}
\def\bTheta{{\mathrm{\Theta}}}

\def\bPsi{{\mathrm{\Psi}}}
\def\nn{\nonumber}
\def\iid{\stackrel{iid}{\sim}}
\def\ind{\stackrel{ind}{\sim}}
\def\hV{\hat{V}}

\def\halpha{\hat{\alpha}}
\def\hbeta{\hat{\beta}}

\def\({\left(}
\def\){\right)}
\def\[{\left[}
\def\]{\right]}

\def\bbE{\mathbb{E}}
\def\bbH{\mathbb{H}}
\def\bbP{\mathbb{P}}
\def\e{\mathrm{e}}
\def\iid{\stackrel{iid}{\sim}}
\def\ind{\stackrel{ind}{\sim}}
\def\nn{\nonumber}
\def\I{\mathbb{I}}

\def\bA{{\mathrm{A}}}

\def\hell{\hat{\ell}}
\def\hu{\hat{u}}

\begin{document}

\begin{frontmatter}

\title{The Discrete Infinite Logistic Normal Distribution}
\runtitle{The Discrete Infinite Logistic Normal Distribution}

\begin{aug}
\author{\fnms{John} \snm{Paisley}},
\author{\fnms{Chong} \snm{Wang}}
\and
\author{\fnms{David} \snm{Blei}}
\runauthor{Paisley, Wang and Blei}
\address{Department of Computer Science\\ Princeton University, Princeton, NJ, USA\\ \{jpaisley,chongw,blei\}@princeton.edu}
\end{aug}

\begin{abstract}
  We present the \textit{discrete infinite logistic normal}
  distribution (DILN), a Bayesian nonparametric prior for mixed
  membership models. DILN generalizes the hierarchical Dirichlet
  process (HDP) to model correlation structure between the weights of
  the atoms at the group level.  We derive a representation of DILN as
  a normalized collection of gamma-distributed random variables and
  study its statistical properties.  We derive a variational inference
  algorithm for approximate posterior inference.  We apply DILN to
  topic modeling of documents and study its empirical performance on
  four corpora, comparing performance with the HDP and the correlated
  topic model (CTM). To compute with large-scale data, we also develop
  a stochastic variational inference algorithm for DILN and compare
  with similar algorithms for HDP and LDA on a collection of $350,000$
  articles from \emph{Nature}.
\end{abstract}

\end{frontmatter}

\section{Introduction}
The hierarchical Dirichlet process (HDP) has emerged as a powerful
Bayesian nonparametric prior for grouped data \citep{Teh:2006},
particularly in its role in Bayesian nonparametric mixed-membership
models.  In an HDP mixed-membership model, each group of data is
modeled with a mixture where the mixture proportions are
group-specific and the mixture components are shared across the data.
While finite models require the number of mixture components to be
fixed in advance, the HDP model allows the data to determine how many
components are needed. And that number is variable: With an HDP model,
new data can induce new components.

The HDP mixed-membership model has been widely applied to
\textit{probabilistic topic modeling}, where hierarchical Bayesian
models are used to analyze large corpora of documents in the service
of exploring, searching, and making predictions about
them~\citep{Blei:2003b,Erosheva:2004,Griffiths:2004a,Blei:2007,Blei:2009}.
In topic modeling, documents are grouped data---each document is a group of
observed words---and we analyze the documents with a mixed-membership
model.  Conditioned on a collection, the posterior expectation of the
mixture components are called ``topics'' because they tend to resemble
the themes that pervade the documents; the posterior expectation of
the mixture proportions identify how each document exhibits the
topics.  Bayesian nonparametric topic modeling uses an HDP to try to
solve the model selection problem; the the number of topics is
determined by the data and new documents can exhibit new topics.

For example, consider using a topic model to analyze 10,000 articles
from Wikipedia.  (This is a data set that we will return to.)  At the
corpus level, the posterior of one component might place high
probability on terms associated with elections; another might place
high probability on terms associated with the military.  At the
document level, articles that discuss both subjects will have
posterior proportions that place weight on both topics.  The posterior
of these quantities over the whole corpus can be used to organize and
summarize Wikipedia in a way that is not otherwise readily available.

Though powerful, the HDP mixed-membership model is limited in that it
does not explicitly model the correlations between the mixing
proportions of any two components.  For example, the HDP topic model
cannot capture that the presence of the election topic in a document
is more positively correlated with the presence of the military topic
than it is a topic about mathematics.  Capturing such patterns, i.e.,
representing that one topic might often co-occur with another, can
provide richer exploratory variables to summarize the data and further
improve prediction.

To address this, we developed the \textit{discrete infinite logistic
  normal distribution} (DILN, pronounced ``Dylan''), a Bayesian
nonparametric prior for mixed-membership
models~\citep{Paisley:2011}.\footnote{In this paper we expand on the
  ideas of \cite{Paisley:2011}, which is a short conference paper.  We
  report on new data analysis, we describe a model of the latent
  component locations that allows for variational inference, we
  improve the variational inference algorithm (see Section
  \ref{sec:kernel}), and we expand it to scale up to very large data
  sets.}  As with the HDP, DILN generates discrete probability
distributions on an infinite set of components, where the same
components are shared across groups but have differently probabilities
within each group.  Unlike the HDP, DILN also models the correlation structure
between the probabilities of the components.

\begin{figure}[h!]\vspace{1cm}
\begin{tabular}{cc}
 \includegraphics[width=.465\textwidth]{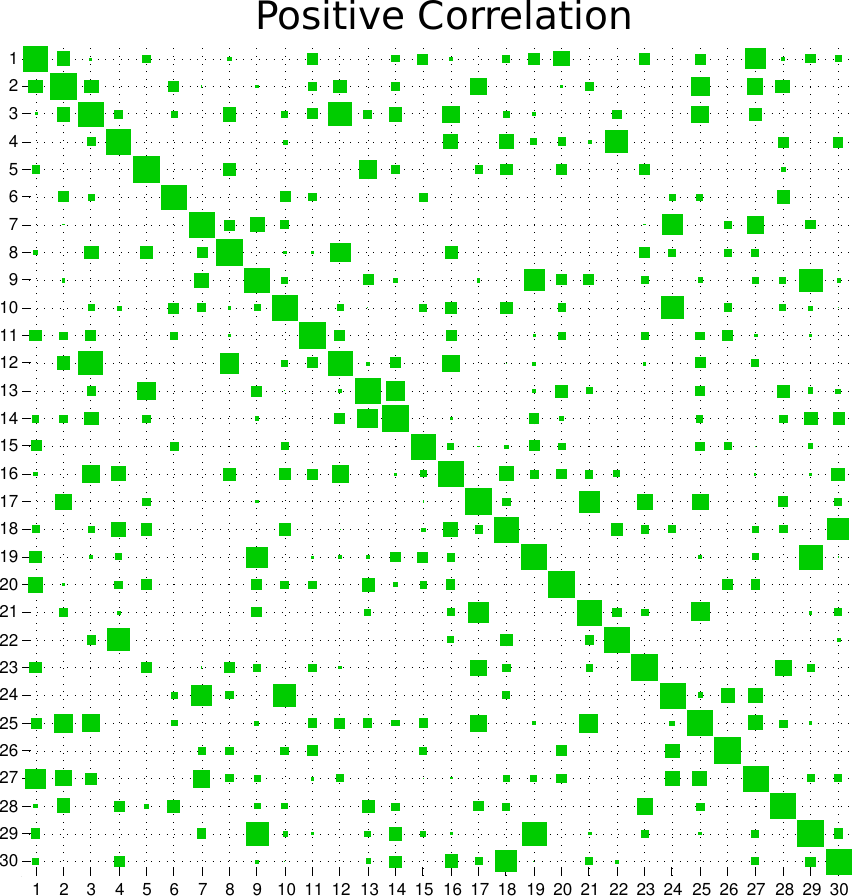}& \includegraphics[width=.465\textwidth]{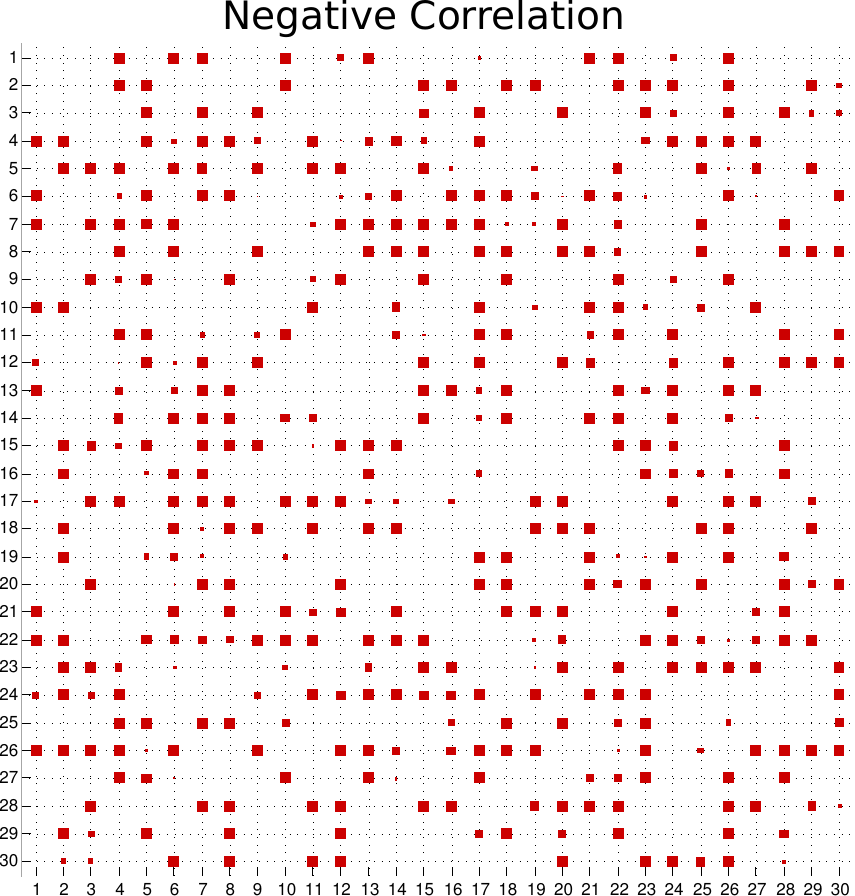}\\
\end{tabular}\vspace{2mm}
\begin{scriptsize}
\begin{tabular}{l}\hline\vspace{-2mm}\\ 
Topic 1: economy, economic, growth, industry, sector, rate, export, production, million, billion\\
Topic 2: international, nations, republic, agreement, relation, foreign, union, nation, china, economic\\
Topic 3: party, election, vote, elect, president, democratic, political, win, minister, seat\\
Topic 4: season, team, win, league, game, championship, align, football, stadium, record\\
Topic 5: treatment, patient, disease, drug, medical, health, effect, risk, treat, symptom\\
Topic 6: album, music, band, record, song, rock, release, artist, recording, label\\
Topic 7: philosophy, philosopher, thing, argument, philosophical, mind, true, truth, reason, existence\\
Topic 8: law, court, legal, criminal, person, rule, jurisdiction, judge, crime, rights\\
Topic 9: math, define, function, theorem, element, definition, space, property, theory, sub\\
Topic 10: church, christian, christ, jesus, catholic, roman, john, god, orthodox, testament\\
Topic 11: climate, mountain, land, temperature, range, region, dry, south, forest, zone\\
Topic 12: constitution, parliament, council, appoint, assembly, minister, head, legislative, house\\
Topic 13: cell, protein, acid, molecule, structure, process, enzyme, dna, membrane, bind\\
Topic 14: atom, element, chemical, atomic, electron, energy, hydrogen, reaction, sup, sub\\
Topic 15: computer, memory, processor, design, hardware, machine, unit, chip, ibm, drive\\
Topic 16: president, congress, washington, governor, republican, john, george, federal, senator, senate\\
Topic 17: military, army, air, unit, defense, navy, service, operation, armed, personnel\\
Topic 18: university, student, school, education, college, program, degree, institution, science, graduate\\
Topic 19: math, value, values, measure, equal, calculate, probability, define, distribution, function\\
Topic 20: food, meat, drink, fruit, eat, vegetable, water, dish, traditional, ingredient\\
Topic 21: battle, commander, command, army, troop, victory, attack, british, officer, campaign\\
Topic 22: sport, ball, team, score, competition, match, player, rule, tournament, event\\
Topic 23: airport, rail, traffic, road, route, passenger, bus, service, transportation, transport\\
Topic 24: religion, god, spiritual, religious, belief, teaching, divine, spirit, soul, human\\
Topic 25: coup, army, military, leader, overthrow, afghanistan, armed, kill, rebel, regime\\
Topic 26: god, goddess, greek, kill, myth, woman, story, sacrifice, ancient, away\\
Topic 27: economic, political, argue, society, social, revolution, free, economics, individual, capitalism\\
Topic 28: radio, service, television, network, station, broadcast, telephone, internet, channel, mobile\\
Topic 29: equation, math, linear, constant, coordinate, differential, plane, frac, solution, right\\
Topic 30: university, professor, prize, award, nobel, research, publish, prise, science, society\\\hline
\end{tabular}
\end{scriptsize}
\caption{Topic correlation for a 10K document Wikipedia corpus: The ten most
  probable words from the 30 most probable topics. At top are the positive and
  negative correlation coefficients for these topics (separated for clarity) as
  learned by the topic locations (see text for details).}\label{fig.wiki}
\end{figure}
Figure~\ref{fig.wiki} illustrates the DILN posterior for 10,000
articles from Wikipedia. The corpus is described by a set of
topics---each topic is a distribution over words and is visualized by
listing the most probable words---and the topics exhibit a correlation
structure.  For example, topic 3 (``party, election, vote'') is
correlated with topic 12 (``constitution, parliament, council'') and
topic 25 (``coup, army, military'').  It is negatively correlated with
topic 20 (``food, meat, drink'').

In DILN, each component is associated with a parameter (e.g., a
topical distribution over terms) and a location in a latent space.
For group-level distributions (e.g., document-specific distributions
over topics), the correlation between component weights is determined
by a kernel function of latent locations of these components.
Since the correlation between occurrences is a posterior correlation,
i.e., one that emerges from the data, the locations of the components
are also latent.  For example, we do not enforce a priori what the
topics are and how they are correlated---this structure comes from the
posterior analysis of the text.

We formulate two equivalent representations of DILN.  We first
formulate it as an HDP scaled by a Gaussian
process~\citep{Rasmussen:2006}.  This gives an intuitive picture of
how the correlation between component weights enters the distribution
and makes clear the relationship between DILN and the HDP.  We then
formulate DILN as a member of the normalized gamma family of random
probability distributions. This lets us characterize the \emph{a
  priori} correlation structure of the component proportions.

The central computational problem for DILN is approximate posterior
inference. Given a corpus, we want to compute the posterior
distribution of the topics, per-document topic proportions, and the
latent locations of the topics.  Using normalized the gamma construction of a random measure, we
derive a variational inference algorithm \citep{Jordan:1999} to
approximate the full posterior of a DILN mixed-membership model.
(Moreover, this variational algorithm can be modified into a new
posterior inference algorithm for HDP mixed-membership models.)  We
use variational inference to analyze several collections of
documents, each on the order of thousands of articles, determining the
number of topics based on the data and identifying an explicit
correlation structure among the discovered topics.  On four
corpora (collected from \textit{Wikipedia}, \textit{Science},
\textit{The New York Times}, and \textit{The Huffington Post}), we
demonstrate that DILN provides a better predictive model and an
effective new method for summarizing and exploring text data. (Again,
see Figure~\ref{fig.wiki} and also Figures \ref{fig.nyt},
\ref{fig.huff} and \ref{fig.sci}.)

Variational inference turns the problem of approximating the posterior
into an optimization problem.  Recent research has used stochastic
optimization to scale variational inference up to very large data
sets~\citep{Hoffman:2010, Dunson:2011}, including our own research on
HDP mixed-membership models~\citep{Wang:2011}.  We used the same
strategy here to develop a scalable inference algorithm for DILN.
This further expands the scope of stochastic variational inference to
models (like DILN) whose latent variables do not enjoy pair-wise
conjugacy.  Using stochastic inference, we analyzed 352,549 thousand
articles from \textit{Nature} magazine, a corpus which would be
computationally expensive with our previous variational algorithm.

\paragraph{Related research.}

The parametric model most closely related to DILN is the
correlated topic model (CTM)~\citep{Blei:2007}.  The CTM is a
mixed-membership model that allows topic occurrences to exhibit
correlation.  The CTM replaces the Dirichlet prior over topic
proportions, which assumes near independence of the components, with a
logistic normal prior~\citep{Aitchison:1982}.  Logistic normal vectors
are generated by exponentiating a multivariate Gaussian vector and
normalizing to form a probability vector.  The covariance matrix of
the multivariate Gaussian distribution provides a means for capturing
correlation structure between topic probabilities.  Our goal in
developing DILN was to form a Bayesian nonparametric variant of this
kind of model.

The natural nonparametric extension of the logistic normal is a
normalized exponentiated Gaussian process
\citep{Lenk:1988,Rasmussen:2006}. However, this cannot function as a
prior for nonparametric correlated topic modeling.  The key property
of the HDP (and DILN) is that the same set of components are shared
among the groups.  This sharing arises because the group-level
distributions on the infinite topic space are discrete probability
measures over the same set of atoms.  Using the model
of~\cite{Lenk:1988} in a hierarchical setting does not provide such
distributions.  The ``infinite CTM'' is therefore not a viable
alternative to the HDP.

In the Bayesian nonparametric literature, another
related line of work focuses on dependent probability distributions
where the dependence is defined on predictors observed for each data
point.  \cite{MacEachern:1999} introduced dependent Dirichlet
processes (DDPs), which allow data-dependent variation in the
atoms of the mixture, and have been applied to spatial modeling
\citep{Gelfand:2005,Rao:2009}.  Other dependent priors allow the
mixing weights themselves to vary with predictors
\citep{Griffin:2006,Duan:2007,Dunson:2008,Ren:2011}.  Still other methods
consider the weighting of multiple DP mixture models using spatial
information~\citep{Muller:2004,Dunson:2007}. 

These methods all use the spatial dependence between observations to
construct observation-specific probability distributions.  Thus they
condition on known locations (often geospatial) for the data.  In
contrast, the latent locations of each component in DILN do not directly
interact with the data, but with each other.  That
is, the correlations induced by these latent locations influence the
mixing weights for a data group \textit{prior} to producing its
observations in the generative process.  Unlike DDP models, our
observations are not equipped with locations and do not \textit{a
  priori} influence component probabilities.  The modeling ideas
behind DILN and behind DDPs are separate, though it is possible to develop
dependent DILN models, just as dependent HDP models have been
developed \citep{MacEachern:1999}.

\paragraph{} This paper is organized as follows.  In Section 2 we
review the HDP and discuss its representation as a normalized gamma
process.  In Section 3 we present the discrete infinite logistic
normal distribution, first as a scaling of an HDP with an
exponentiated Gaussian process and then using a normalized gamma
construction.  In Section 4 we use this gamma construction
to derive a mean-field variational inference algorithm for approximate
posterior inference of DILN topic models, and we extend this algorithm
to the stochastic variational inference setting.  Finally, in Section
5 we provide an empirical study of the DILN topic model on five text
corpora.

\section{Background: The Hierarchical Dirichlet Process}

The discrete infinite logistic normal (DILN) prior for mixed-membership models is an extension of the hierarchical Dirichlet process (HDP)~\citep{Teh:2006}.  In this
section, we review the HDP and reformulate it as a normalized gamma process.

\subsection{The original formulation of the hierarchical Dirichlet process}
The Dirichlet process \citep{Ferguson:1973} is useful as a Bayesian nonparametric prior for mixture models since it generates distributions on infinite parameter spaces that are almost surely discrete \citep{Blackwell:1973,Sethuraman:1994}. Given a space $\Omega$ with a corresponding Borel $\sigma$-algebra $\mathcal{B}$ and base measure $\alpha G_0$, where $\alpha > 0$ and $G_0$ is a probability measure, \cite{Ferguson:1973} proved the existence of a process $G$ on $(\Omega,\mathcal{B})$ such that for all measurable partitions $\{B_1,\dots,B_K\}$ of $\Omega$,
\begin{equation}
 (G(B_1),\dots,G(B_K)) \sim \mathrm{Dirichlet}(\alpha G_0(B_1),\dots,\alpha G_0(B_K)).
\end{equation}
This is called a Dirichlet process and is denoted $G \sim \DP(\alpha G_0)$. \cite{Sethuraman:1994} gave a proof of the almost sure (a.s.) discreteness of $G$ by way of a stick-breaking representation \citep{Ishwaran:2001}; we will review this stick-breaking construction later. \cite{Blackwell:1973} gave an earlier proof of this discreteness using P\'{o}lya urn schemes. The discreteness of $G$ allows us to write it as
\begin{equation*}
  G = \sum_{k=1}^{\infty} p_k\delta_{\eta_k},
\end{equation*}
where each atom $\eta_k$ is generated i.i.d.\ from the base distribution $G_0$, and the
atoms are given random probabilities $p_k$ whose distribution depends on a
scaling parameter $\alpha > 0$ such that smaller values of $\alpha$ lead to
distributions that place more mass on fewer atoms.  The DP is most commonly
used as a prior for a mixture model, where $G_0$ is a distribution on a model parameter, $G\sim \DP(\alpha G_0)$ and each data
point is drawn from a distribution family indexed by a parameter drawn from $G$~\citep{Ferguson:1983,Lo:1984}.

When the base measure $G_0$ is non-atomic, multiple draws from the DP prior place their probability mass on an a.s.\ disjoint set of atoms. That is, for $G_1,G_2 \iid \DP(\alpha G_0)$, an atom $\eta_k$ in $G_1$ will a.s.\ not appear in $G_2$, i.e., $G_1(\{\eta_k\}) > 0 \implies G_2(\{\eta_k\}) = 0$ a.s. The goal of mixed-membership modeling is to use all groups of data to learn a shared set of atoms. The
hierarchical Dirichlet process~\citep{Teh:2006} was introduced to allow multiple
Dirichlet processes to share the same atoms. The HDP is a prior for a collection of random distributions $(G_1,\dots,G_M)$. Each $G_m$ is i.i.d.\ DP distributed with a base probability measure that is also a Dirichlet process,
\begin{equation}
G \sim \mathrm{DP}(\alpha G_0), \quad\quad
G_m\,|\,G ~ \iid ~ \mathrm{DP}(\beta G).
\end{equation}
The hierarchical structure of the HDP ensures that each $G_m$ has probability
mass distributed across a shared set of atoms, which results from the a.s.\ discreteness of the second-level base measure $\beta G$. Therefore, the same subset of atoms
will be used by all groups of data, but with different probability
distributions on these atoms for each group.

Where the DP allows us to define a mixture model, the HDP allows us to define a
mixed-membership model. Given an HDP $(G_1,\dots,G_M)$, each $G_m$ generates its associated group of data from a mixture model,
\begin{eqnarray}
  X_n^{(m)}\,|\,\theta_n^{(m)} & \ind & f(X|\theta_n^{(m)}),\quad n=1,\dots,N_m,\\
  \theta_n^{(m)}\,|\,G_m & \iid & G_m,\quad\quad\quad\quad n = 1,\dots,N_m.
\end{eqnarray}
The datum $X_n^{(m)}$ denotes the $n$th observation in the $m$th group and $\theta_n^{(m)}$ denotes its associated parameter drawn from the mixing distribution $G_m$, with $\mathrm{Pr}(\theta_n^{(m)} = \eta_k|G_m) = G_m(\{\eta_k\})$.  The
HDP can be defined to an arbitrary depth, but we focus on the
two-level process described above.

When used to model documents, the HDP is a prior for topic models.  The
observation $X_n^{(m)}$ is the $n$th word in the $m$th document and is drawn from a discrete distribution on words in a vocabulary, $X_n^{(m)}|\theta_n^{(m)} \sim \mathrm{Discrete}(\theta_n^{(m)})$, where $\theta_n^{(m)}$ is the $V$-dimensional word probability vector selected according to $G_m$ by its corresponding word. The base probability measure $G_0$ is usually a symmetric Dirichlet distribution on the vocabulary simplex.  Given a document collection, posterior inference yields a
set of shared topics and per-document proportions over all topics. Unlike its
finite counterpart, latent Dirichlet allocation~\citep{Blei:2003b}, the HDP
topic model determines the number of topics from the data~\citep{Teh:2006}.

\subsection{The HDP as a normalized gamma process}\label{sec:HDP}

The DP has several representations, including a gamma process
representation \citep{Ferguson:1973} and a stick-breaking
representation \citep{Sethuraman:1994}.  In constructing HDPs, we
will take advantage of each of these representations at different
levels of the hierarchy.

We construct the top-level DP using stick-breaking~\citep{Sethuraman:1994},
\begin{equation}\label{eqn.sethuraman}
G = \sum_{k=1}^{\infty} V_k\prod_{j=1}^{k-1}(1-V_{j})\delta_{\eta_k}, \quad V_k \iid \mathrm{Beta}(1,\alpha), \quad \eta_k \iid G_0.
\end{equation}
The name comes from an interpretation of $V_k$ as the proportion broken
from the remainder of a unit-length stick $\prod_{j=1}^{k-1}(1-V_j)$.
The resulting absolute length of this stick forms the probability of
atom $\eta_k$. Letting $p_k := V_k\prod_{j=1}^{k-1}(1-V_{j})$, this
method of generating DPs produces probability measures that are
size-biased according to index $k$ since $\mathbb{E}[p_k] >
\mathbb{E}[p_j]$ for $k < j$.

Turning to the second-level DP $G_m$, we now use a normalized gamma
process. Recall that a $K$-dimensional Dirichlet-distributed vector
$(Y_1,\dots,Y_K) \sim \mathrm{Dirichlet}(c_1,\dots,c_K)$ with $c_i >
0$ and $\sum_j c_j < \infty$ can be generated for any value of $K$ by
drawing $Z_i \ind \mathrm{Gamma}(c_i,1)$ and defining $Y_i :=
Z_i/\sum_j Z_j$ \citep{Ishwaran:2002}. \cite{Ferguson:1973} focused on
the infinite extension of this representation as a normalized gamma
process.  Since $p_k > 0$ for all atoms $\eta_k$ in $G$, and also
because $\sum_{j=1}^{\infty} \beta p_j = \beta < \infty$, we can
construct each $G_m$ using the following normalization of a gamma
process,
\begin{equation}\label{eqn.gammaHDP}
G_m\,|\, G,Z = \sum_{k=1}^{\infty} \frac{Z_k^{(m)}}{\sum_{j=1}^{\infty}Z_{j}^{(m)}}\delta_{\eta_k}, \quad Z_k^{(m)}\,|\, G \ind \mathrm{Gamma}(\beta p_k,1).
\end{equation}
The gamma process representation of the DP is discussed
by~\cite{Ferguson:1973}, \cite{Kingman:1993} and \cite{Ishwaran:2002},
but it has not been applied to the HDP.  In DILN we will mirror this
type of construction of the HDP---a stick-breaking construction for
the top-level DP and a gamma process construction for the
second-level DPs.  This will let us better articulate model properties and also
make inference easier.

\section{The Discrete Infinite Logistic Normal Distribution}

The HDP prior has the hidden assumption that the presence of one atom in
a group is not \emph{a priori} correlated with the presence of
another atom (aside from the negative correlation imposed by the
probability simplex). At the group level the HDP cannot model
correlation structure between the components' probability mass.  To
see this, note that the gamma process used to construct each
group-level distribution is an example of a completely random
measure~\citep{Kingman:1993}.  That is, the unnormalized masses
$(Z_1^{(m)}, Z_2^{(m)},\dots)$ of the atoms $(\eta_1,\eta_2,\dots)$ of
$G_m$ are independently drawn, and for all partitions
$\{B_1,\dots,B_K\}$ of $\Omega$ and given $S_m := \sum_j Z_j^{(m)}$,
the scaled random variables $S_mG_m(B_1),\dots,S_mG_m(B_K)$ are
independent.  Thus, no correlation between per-group probabilities can
be built into the HDP.

We introduced the discrete infinite logistic normal (DILN) as a
modification of the HDP that can express such
correlations~\citep{Paisley:2011}.  The idea is that each atom lives
in a latent location, and the correlation between atom probabilities
is determined by their relative locations in the latent space.  When analyzing
data, modeling these correlations can improve the predictive
distribution and provide more information about the underlying latent
structure.  DILN has two equivalent representations; we first describe
it as a scaled HDP, with scaling determined by an exponentiated
Gaussian process~\citep{Rasmussen:2006}. We then show how DILN fits
naturally within the family of normalized gamma constructions of
discrete probability distributions in a way similar to the discussion
in Section \ref{sec:HDP} for the HDP.

\subsection{DILN as a scaled HDP}\label{sec.DILNasHDP}
DILN shares the same hierarchical structure described in Section
\ref{sec:HDP} for the HDP---there is an infinite set of components and
each group exhibits those components with different probabilities.  In
DILN, we further associate each component with a latent location in
$\realline^d$.  (The dimension $d$ is predefined.)  The model then
uses these locations to influence the correlations between the
probabilities of the components for each group-level distribution.  In
posterior inference, we infer both the components and their latent
locations.  Thus, through the inferred locations, we can estimate the
correlation structure among the components.

Let $G_0$ be a base distribution over parameter values $\eta \in
\Omega$, and let $L_0$ be a non-atomic base distribution over
locations, $\ell \in \realline^d$. We first draw a top-level Dirichlet
process with a product base measure $\alpha G_0\times L_0$,
\begin{equation}\label{eqn.toplevel}
  G \sim \textrm{DP}(\alpha G_0 \times L_0).
\end{equation}
Here, $G$ is a probability measure on the space $\Omega \times \realline^d$. For each atom $\{\eta,\ell\} \in \Omega \times \realline^d$, we think of $\eta\in\Omega$ as living in the parameter space, and $\ell\in\realline^d$ as living in the location space.

In the second level of the process, the model uses both the probability measure
$G$ and the locations of the atoms to construct group-level probability
distributions. This occurs in two steps. In the first step, we independently draw a Dirichlet process and a Gaussian process using the measure and atoms of $G$,  
\begin{equation}
G_m^{\textrm{DP}}\,|\,G \sim \textrm{DP}(\beta G),\quad W_m(\ell) \sim \textrm{GP}(\boldsymbol{\mu}(\ell), \bK(\ell, \ell')).
\end{equation}
The Dirichlet process $G_m^{\DP}$ provides a new, initial distribution on the atoms of $G$ for group $m$. The Gaussian process $W_m$ is defined on the locations of the atoms of $G$ and results in a random function that can be evaluated using the location of each atom.
The covariance between $W_m(\ell)$ and $W_m(\ell')$ is determined by a
kernel function $\bK(\ell, \ell')$ on their respective locations.

The second step is to form each group-level distribution by scaling the probabilities of each second-level
Dirichlet process by the exponentiated values of its corresponding Gaussian process,
\begin{equation}\label{eqn.scaledHDP}
  G_m(\{\eta,\ell\})\,|\, G_m^{\DP},W_m \propto G_m^{\textrm{DP}}(\{\eta,\ell\}) \exp\{W_m(\ell)\}.
\end{equation}
Since we define $G_0$ and $L_0$ to be non-atomic, all $\eta$ and
$\ell$ in $G$ are a.s. distinct, and evaluating the Gaussian process
$W_m$ at a location $\ell$ determines its atom $\{\eta,\ell\}$. We satisfy two objectives with this representation: ($i$) the
probability measure $G_m$ is discrete, owing to the discreteness of
$G_m^{\textrm{DP}}$, and ($ii$) the probabilities in $G_m$ are
\textit{explicitly} correlated, due to the exponentiated Gaussian
process. We emphasize that these correlations arise from latent
locations and in posterior inference we infer these locations from
data. 

\begin{figure}[t] \centering
  \includegraphics[width=.98\textwidth]{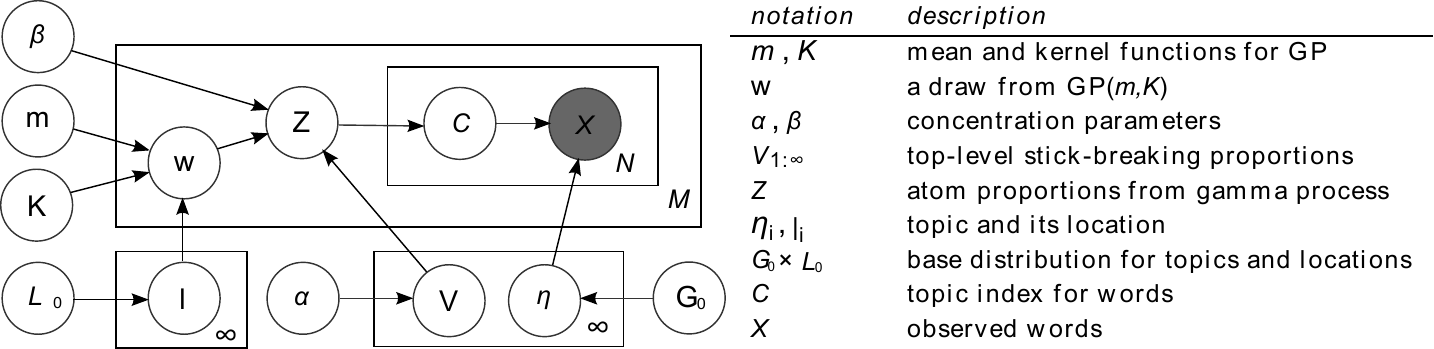}
  \caption{A graphical model of the normalized gamma construction  of the
    DILN topic model.} \label{fig.graphicalmodel}
\end{figure}
\subsection{A normalized gamma construction of DILN}
\label{sec:iln-construction}

We now turn to a normalized gamma construction of DILN.  We show
that the DILN prior uses the second parameter of the gamma
distribution in the normalized gamma construction of the HDP to
model the covariance structure among the components of $G_m$. This
representation facilitates approximate posterior inference described
in Section~\ref{sec:vb}, and helps clarify the covariance properties
of the group-level distributions over atoms.

We use a stick-breaking construction of the top-level Dirichlet
process (Equation \ref{eqn.toplevel}),
\begin{equation}\label{eqn.topleveliLN}
  G = \sum_{k=1}^{\infty}
  V_k\prod_{j=1}^{k-1}(1-V_{j})\delta_{\{\eta_k,\ell_k\}}, \quad V_k
  \stackrel{iid}{\sim} \mathrm{Beta}(1,\alpha), \quad \eta_k
  \stackrel{iid}{\sim} G_0, \quad \ell_k \stackrel{iid}{\sim} L_0.
\end{equation}
This is nearly the same as the top-level construction of the HDP given in Equation
(\ref{eqn.sethuraman}). The difference is that the product base measure
is defined over the latent location $\ell_k$ as well as the component
$\eta_k$ to form the atom $\{\eta_k,\ell_k\}$.

We pattern the group-level distributions after the gamma process
construction of the second-level DP in the HDP,
\begin{equation}
  \label{eqn.secondleveliLN2}
  G_m\,|\, G,Z = \sum_{k=1}^{\infty} \frac{Z_k^{(m)}}{\sum_{j=1}^{\infty}
    Z_{j}^{(m)}} \delta_{\{\eta_k,\ell_k\}},
\end{equation}
\begin{equation} \nonumber
\quad Z_k^{(m)}\,|\, G,W_m \sim \mathrm{Gamma}(\beta p_k,\exp\{-W_m(\ell_k)\}), \quad W_m\,|\, G \stackrel{iid}{\sim} \mathrm{GP}(\boldsymbol{\mu}(\ell),\bK(\ell,\ell')),
\end{equation}
with $p_k := V_k\prod_{j=1}^{k-1}(1-V_{j})$.  Here, DILN differs from the HDP in
that it uses the second parameter of the gamma distribution.  In the
appendix, we give a proof that the normalizing constant is almost
surely finite.

We note that the locations $\ell_k$ contained in each atom no longer
serve a function in the model \textit{after} $G_m$ is constructed, but
we include them in Equation (\ref{eqn.secondleveliLN2}) to be technically
correct. The purpose of the locations $\{\ell_k\}$ is to generate
sequences $Z_1^{(m)},Z_2^{(m)},\dots$ that are correlated, which is
not achieved by the HDP. After constructing the weights of $G_m$, the
locations have fulfilled their role and are no longer used downstream
by the model.

We derive Equation (\ref{eqn.secondleveliLN2})
using a basic property of gamma distributed random variables.  Recall
that the gamma density is $f(z|a,b) = b^a
z^{a-1}\exp\{-bz\}/\Gamma(a)$.  Consider a random variable $y \sim
\mathrm{Gamma}(a,1)$ that is scaled by $b > 0$ to produce $z = b y$.
Then $z \sim \mathrm{Gamma}(a,b^{-1})$.  In Equation
(\ref{eqn.scaledHDP}) we scale atom $\{\eta,\ell\}$ of the Dirichlet
process $G_m^{\DP}$ by $\exp\{W_m(\ell)\}$. Using the gamma process
representation of $G_m^{\DP}$ given in Equation (\ref{eqn.gammaHDP}) and
the countably infinite $G$ in Equation (\ref{eqn.topleveliLN}), we have
that $G_m(\{\eta_k,\ell_k\}) \propto Y_k^{(m)}\exp\{W_m(\ell)\}$,
where $Y_k^{(m)} \sim \mathrm{Gamma}(\beta p_k,1)$. Since $Z_k^{(m)}
:= Y_k^{(m)}\exp\{W_m(\ell)\}$ is distributed as $\mathrm{Gamma}(\beta
p_k,\exp\{-W_m(\ell_k)\})$ by the above property of scaled gamma
random variables, the construction in Equation
(\ref{eqn.secondleveliLN2}) follows.

For the topic model, drawing an observation proceeds as for the HDP. We use a latent indicator variable $C_n^{(m)}$, which selects the index of the atom used by observation $X_n^{(m)}$. This indicator variable gives a useful hidden-data representation of the process for inference in mixture models \citep{Escobar:1995},
\begin{equation}
  X_n^{(m)}\,|\, G_m, C_n^{(m)} \ind \mathrm{Discrete}(\eta_{C_n^{(m)}}), \quad  C_n^{(m)}\,|\, G_m \iid \sum_{k=1}^{\infty} \frac{Z_k^{(m)}}{\sum_{j=1}^{\infty} Z_{j}^{(m)}} \delta_k\, ,
\end{equation}
where the discrete distribution is on word index values $\{1,\dots,V\}$. We note that this discrete distribution is one of many possible data generating
distributions, and changing this distribution and $G_0$ will allow for DILN to be used in a variety of other mixed-membership modeling applications \citep{Erosheva:2007,Airoldi:2008,Pritchard:2000}.
Figure~\ref{fig.graphicalmodel} shows the graphical model of the
DILN topic model.

\subsection{The covariance structure of DILN}

The two-parameter gamma representation of DILN permits simple calculation of the
expectation, variance and covariance prior to normalization. We first give these values conditioning on the top-level Dirichlet process $G$ and integrating out the Gaussian process $W_m$. In the following calculations, we assume that the mean function of the Gaussian process is $\boldsymbol{\mu}(\cdot) = 0$ and we define $k_{ij} := \bK(\ell_i,\ell_j)$. The expectation, variance and
covariance of $Z_i^{(m)}$ and $Z_j^{(m)}$ are
\begin{eqnarray}\label{eqn.expvar}
\mathbb{E}\left[Z^{(m)}_i|\beta, \boldsymbol{p}, \bK\right] \hspace{-.2cm}&=&\hspace{-.2cm} \beta p_i \e^{\frac{1}{2}k_{ii}},\\
\mathbb{V}\left[Z^{(m)}_i|\beta, \boldsymbol{p}, \bK\right] \hspace{-.2cm}&=&\hspace{-.2cm} \beta p_i \e^{2k_{ii}} + \beta^2 p_i^2 \e^{k_{ii}}\(\e^{k_{ii}} - 1\),\nn\\
\mathrm{Cov}\left[Z^{(m)}_i,Z^{(m)}_j|\beta, \boldsymbol{p}, \bK\right] \hspace{-.2cm}&=&\hspace{-.2cm} \beta^2 p_i p_j \e^{\frac{1}{2}(k_{ii}+k_{jj})}\left(\e^{k_{ij}} - 1\right).\nn
\end{eqnarray}
Observe that the covariance is similar to the unnormalized logistic
normal~\citep{Aitchison:1982}, but with the additional term $\beta^2
p_i p_j$. In general, these $p_i$ terms show how sparsity is enforced
by the top-level DP, since both the expectation and variance terms go to
zero exponentially fast as $i$ increases.

These values can also be calculated with the top-level Dirichlet process integrated out using the tower property of conditional expectation. They are
\begin{eqnarray}\label{eqn.moments2}
  \mathbb{E}\left[Z^{(m)}_i|\alpha, \beta, \bK\right] &=& \beta \mathbb{E}[p_i] \e^{\frac{1}{2}k_{ii}},\\
  \mathbb{V}\left[Z^{(m)}_i|\alpha, \beta, \bK\right] &=& \beta \mathbb{E}[p_i] \e^{2k_{ii}} + \beta^2 \mathbb{E}[p_i^2] \e^{2k_{ii}} - \beta^2 \mathbb{E}[p_i]^2 \e^{k_{ii}},\nonumber\\
  \mathrm{Cov}\left[Z^{(m)}_i,Z^{(m)}_j|\alpha, \beta, \bK\right] &=& \beta^2 \mathbb{E}[p_ip_j] \e^{\frac{1}{2}(k_{ii}+k_{jj}) + k_{ij}} - \beta^2\mathbb{E}[p_i]\mathbb{E}[p_j]\e^{\frac{1}{2}(k_{ii}+k_{jj})}.\nonumber
\end{eqnarray}
The values of the expectations in Equation (\ref{eqn.moments2}) are
\begin{equation}\nonumber
 \mathbb{E}[p_i] = \frac{\alpha^{i-1}}{(1+\alpha)^i},\quad \mathbb{E}[p_i^2] = \frac{2\alpha^{i-1}}{(1+\alpha)(2+\alpha)^i},\quad \mathbb{E}[p_ip_j] = \frac{\alpha^{i-1}}{(2+\alpha)^j(1+\alpha)^{i-j+1}}, ~~ i > j.
\end{equation}
Note that some covariance remains when $k_{ij} = 0$, since the conditional
independence induced by $\boldsymbol{p}$ is no longer present.  The available
covariance structure depends on the kernel. For example, when a Gaussian kernel
is used, a structured negative covariance is not achievable since $k_{ij} \geq 0$. We next
discuss one possible kernel function, which we will use in our
inference algorithm and experiments.

\subsection{Learning the kernel for DILN}
\label{sec:kernel}

In our formulation of DILN, we have left the kernel function
undefined.  In principle, any kernel function can be used, but in
practice some kernels yield simpler inference algorithms than others.
For example, while a natural choice for $\bK(\ell,\ell')$ is the
Gaussian kernel, we found that the resulting variational inference
algorithm was computationally expensive because it required many
matrix inversions to infer the latent locations.\footnote{In
  \cite{Paisley:2011} we side-stepped this issue by learning a point
  estimate of the matrix $\bK$, which was finite following a truncated
  approximation introduced for variational inference. We suggested
  finding locations by using an eigendecomposition of the learned
  $\bK$.  The approach outlined here is more rigorous in that it stays
  closer to the model and is not tied to a particular approximate
  inference approach.}  In this section, we define an alternative
kernel.  In the next section, we will see that this leads to simple
algorithms for approximate inference of the latent locations $\ell$.

We model the location of a component with a zero-mean Gaussian vector
in $\mathbb{R}^d$. We then form the kernel by taking the dot product
of these vectors. That is, for components $k$ and $j$, we draw
locations and parameterize the Gaussian process for $W_m$ as
\begin{equation}\label{eqn.kernel}
  \ell_k \iid \mbox{Normal}(0,c I_d), \quad\quad \boldsymbol{\mu}(\ell_k) = 0,
  \quad\quad \bK(\ell_k,\ell_j) = \ell_k^T\ell_j.
\end{equation}

With this specification, all $p$-dimensional ($p\leq d$) sub-matrices
of $\bK$ are Wishart-distributed with parameters $p$ and $cI_p$
\citep{Dawid:1981}. However, this kernel is problematic.  When the
number of components $p$ is greater than $d$, it will produce singular
covariance matrices that cannot be inverted in the Gaussian likelihood
function of $W_m$, an inversion that is required during inference.
While in parametric models we might place constraints on the number of
components, our prior is nonparametric.  We have an infinite number of
components and therefore $\bK$ must be singular.

We solve this problem by forming an equivalent representation of the
kernel in Equation (\ref{eqn.kernel}) that yields a more tractable joint
likelihood function. This representation uses auxiliary variables as follows.
Let $u\sim \mathrm{Normal}(0,I_d)$ and recall that for a vector $z =
B^Tu$, the marginal distribution of $z$ is $z|B \sim
\mathrm{Normal}(0,B^TB)$. In our case, $B^TB$ is the inner
product kernel and the columns of $B$ correspond to component
locations, $B = [\ell_1, \ell_2, \cdots]$.

With this in mind, we use the following construction of the Gaussian process $W_m$,
\begin{equation}
  \quad W_m(\ell_k) = \ell^T_k u_m, \quad\quad u_m \sim \mathrm{Normal}(0,I_d).
\end{equation}
Marginalizing the auxiliary vector $u_m$ gives the desired $W_m(\ell_k) \sim \mathrm{GP}(0,\bK(\ell_k,\ell_j))$.

The auxiliary vector $u_m$ allows for tractable inference of Gaussian
processes that lie in a low-dimensional subspace. Aside from
analytical tractability, the vector $u_m$ can be interpreted as a
location for group $m$.  (This is not to be confused with the location
of component $k$, $\ell_k$.)  The group locations let us measure
similarity between groups, such as document similarity in the topic
modeling case.  In the following sections, we no longer work directly
with $W_m(\ell_k)$, but rather the dot product $\ell_k^Tu_m$ through
inference of $\ell$ and $u$.

\section{Variational Inference for DILN}\label{sec:vb}

In Bayesian nonparametric mixed-membership modeling, the central
computational problem is posterior inference.  However, computing the
exact posterior is intractable. For HDP-based models, researchers have
developed several approximate methods
\citep{Teh:2006,Liang:2007,Teh:2009,Wang:2011}.

In this paper, we derive a mean-field variational inference algorithm
\citep{Jordan:1999,Wainwright:2008} to approximate the posterior of a
DILN mixed-membership model. We focus on topic modeling but note that
our algorithm can be applied (with a little modification) to any DILN
mixed-membership model.  In addition, since the HDP is an instance of
DILN, this algorithm also provides an inference method for HDP
mixed-membership models.

Variational methods for approximate posterior inference attempt to
minimize the Kullback-Leibler divergence between a factorized
distribution over the hidden variables and the true posterior.  The
hidden variables in the DILN topic model can be broken into
document-level variables (those defined for each document), and
corpus-level variables (those defined across documents); the
document-level variables are the unnormalized weights $Z_k^{(m)}$,
topic indexes $C_n^{(m)}$, and document locations $u_m$; the
corpus-level variables are the topic distributions $\eta_k$,
proportions $V_k$, concentration parameters $\alpha$ and $\beta$, and
topic locations $\ell_k$. Under the mean-field assumption the
variational distribution that approximates the full posterior is
factorized,
\begin{equation}
  Q := q(\alpha)q(\beta)\prod_{k=1}^T q(\eta_k)q(V_k)q(\ell_k)
  \textstyle \prod_{m=1}^M q(Z_k^{(m)})q(C_n^{(m)})q(u_m).
\end{equation}
We select the following variational distributions for each latent
variable,
\begin{eqnarray}
  q(C_n^{(m)}) &=& \mathrm{Multinomial}(C_n^{(m)}|\phi_n^{(m)})\nn\\
  q(Z_k^{(m)}) &=& \mathrm{Gamma}(Z_k^{(m)}|a_k^{(m)},b_k^{(m)})\nn\\
  q(\eta_k)    &=& \mathrm{Dirichlet}(\eta_k|\gamma_{k,1},\dots,\gamma_{k,D})\nn\\
  q(\ell_k)q(u_m) &=& \delta_{\hell_k}\cdot\delta_{\hu_m}\nn\\
  q(V_k) &=& \delta_{\hV_k}\nn\\
  q(\alpha)q(\beta) &=& \delta_{\halpha}\cdot\delta_{\hbeta}\,.
\end{eqnarray}
The set of parameters to these distributions are the \textit{variational
parameters}, represented by $\bPsi$. The goal of variational inference is to optimized these parameters to make
the distribution $Q$ close in KL divergence to the true posterior. Minimizing
this divergence is equivalent to maximizing a lower bound on the log marginal likelihood
obtained from Jensen's inequality,
\begin{equation}\label{eqn.jensens}
 \ln \int p(\bX,\bTheta)\, d\bTheta \geq \int Q(\bPsi)\ln
 \frac{p(\bX,\bTheta)}{Q(\bPsi)}\, d\bTheta,
\end{equation}
where $\bTheta$ stands for all hidden random variables.  This
objective has the form
\begin{equation}
  \label{eqn.jensens2}
 \mathcal{L}(\bX,\bPsi) = \mathbb{E}_Q[\ln p(\bX,\bTheta)] + \mathbb{H}[Q].
\end{equation}
We will find a locally optimal solution of this function using
coordinate ascent, as detailed in the next section.

Note that we truncate the number of components at $T$ in the top-level
Dirichlet process \citep{Blei:2005}. \cite{Kurihara:2006} show how
infinite-dimensional objective functions can be defined for
variational inference, but the conditions for this are not met by
DILN. The truncation level $T$ should be set larger than the total
number of topics expected to be used by the data. A value of $T$ that
is set too small is easy to diagnose: the approximate posterior will
use all $T$ topics. Setting $T$ large enough, the variational
approximation will prefer a corpus-wide distribution on topics that is
sparse. We contrast this with the CTM and other finite topic models,
which fit a pre-specified number of topics to the data and potentially
overfit if that number is too large.

We have selected several delta functions as variational
distributions. In the case of the top-level stick-breaking proportions
$V_k$ and second-level concentration parameter $\beta$, we have
followed \cite{Liang:2007} in doing this for tractability. In the case
of the top-level concentration parameter $\alpha$, and topic and
document locations $\ell_k$ and $u_m$, these choices simplify the
algorithm.

\begin{algorithm}[tb]
   \caption{Batch variational Bayes for DILN}
   \label{alg:alg1}
\raggedright
Batch optimization of the variational lower bound $\mathcal{L}$\\
Optimize corpus-wide and document-specific variational parameters $\bPsi'$ and $\bPsi_m$\\
\begin{algorithmic}[1]
  \WHILE{$\bPsi'$ and $\bPsi_m$ have not converged}
  \FOR{$m = 1,\dots,M$}
  \STATE Optimize $\bPsi_m$ ({Equations \ref{eqn.upC}--\ref{eqn.upU}})
  \ENDFOR
  \STATE Optimize $\bPsi'$ (Equations \ref{eqn.upEta}--\ref{eqn.upBeta})
  \ENDWHILE
\end{algorithmic}
\end{algorithm}

\subsection{Coordinate ascent variational inference}\label{sec:vbalg}

We now present the variational inference algorithm for the DILN topic
model.  We optimize the variational parameters $\bPsi$ with respect to
the variational objective function of Equation (\ref{eqn.jensens2}).
For DILN, the variational objective expands to
\begin{eqnarray}\label{eqn.lowerbound}
\mathcal{L} \hspace{-1mm}&=&\hspace{-1mm} \sum_{m=1}^M\sum_{n=1}^{N_m}\sum_{k=1}^T \phi_{n,k}^{(m)}\mathbb{E}_q[\ln p(X_n^{(m)}|\eta_k)] + \sum_{m=1}^M\sum_{n=1}^{N_m}\sum_{k=1}^T \phi_{n,k}^{(m)}\mathbb{E}_q[\ln p(C_n^{(m)} = k | Z_{1:T}^{(m)})]\nn \\
\hspace{-1mm}& + &\hspace{-1mm} \sum_{m=1}^M\sum_{k=1}^T \mathbb{E}_q[\ln p(Z_k^{(m)}|\beta p_k,\ell_k,u_m)] +\sum_{k=1}^T\mathbb{E}_q[\ln p(\eta_k|\gamma)] + \sum_{k=1}^T \mathbb{E}_q[ \ln p(V_k|\alpha)]\nn\\
\hspace{-1mm}& +&\hspace{-1mm} \sum_{k=1}^T \mathbb{E}_q[\ln p(\ell_k)] +  \sum_{m=1}^M \mathbb{E}_q[\ln p(u_m)] + \mathbb{E}_q[\ln p(\alpha)] + \mathbb{E}_q[\ln p(\beta)]  - \mathbb{E}_Q[\ln Q].
\end{eqnarray}
We use coordinate ascent to optimize this function, iterating between
two steps.  In the first step we optimize the document-level
parameters for each document; in the second step we optimize the
corpus-level parameters. Algorithm \ref{alg:alg1} summarizes this general inference structure.

\subsubsection*{\textbf{Document-level parameters}}

For each document, we iterate between updating the variational distribution of
per-word topic indicators $C_n^{(m)}$, unnormalized weights $Z_k^{(m)}$, and
document locations $\hu_m$.

\paragraph{Coordinate update of $q(C_n^{(m)})$} The variational
distribution on the topic index for word $X_n^{(m)}$ is multinomial
with parameter $\phi$. For $k=1,\dots,T$ topics
\begin{equation}\label{eqn.upC}
  \phi_{n,k}^{(m)}\propto \exp\left\lbrace\E_Q[\ln
    \eta_k(X_n^{(m)})]+\E_Q[\ln Z_k^{(m)}]\right\rbrace.
\end{equation}
Since $\phi_n^{(m)} = \phi_{n'}^{(m)}$ when $X_n^{(m)} =
X_{n'}^{(m)}$, we only need to compute this update once for each
unique word occurring in document $m$.

\paragraph{Coordinate update of $q(Z_k^{(m)})$} This variational gamma
distribution has parameters $a_k^{(m)}$ and $b_k^{(m)}$. Let $N_m$ be
the number of observations (e.g., words) in group $m$.  After
introducing an auxiliary parameter $\xi_m$ for each group-level
distribution (discussed below), the updates are
\begin{eqnarray}\label{eqn.upZ}
a_k^{(m)} & = & \hat{\beta} p_k + \sum_{n=1}^{N_m} \phi_{n,k}^{(m)},\nn\\
b_k^{(m)} & = & \exp\{-\hell_k^T\hu_m\} + \frac{N_m}{\xi_m}.
\end{eqnarray}
We again denote the top-level stick-breaking weights by $p_k =
\hV_k\prod_{j=1}^{k-1}(1-\hV_j)$. The expectations from this
distribution that we use in subsequent updates are $\E_Q[Z_k^{(m)}] =
a_k^{(m)}/b_k^{(m)}$ and $\E_Q[\ln Z_k^{(m)}] = \psi(a_k^{(m)}) - \ln
b_k^{(m)}$.

The auxiliary parameter allows us to approximate the term
$\mathbb{E}_Q[\ln p(C_n^{(m)} = k | Z_{1:T}^{(m)})]$ appearing in the
lower bound. To derive this, we use a first order Taylor expansion on
the following intractable expectation,
\begin{equation}\nonumber
  -\E_Q\[\ln \sum_{k=1}^T Z_k^{(m)}\] \geq - \ln \xi_m -
  \frac{\sum_{k=1}^T\E_Q[Z_k^{(m)}]-\xi_m}{\xi_m}.
\end{equation}
The update for the auxiliary variable $\xi_m$ is $\xi_m = \sum_{k=1}^T
\E_Q[Z_k^{(m)}].$ See the appendix for the complete derivation.

\paragraph{Coordinate update of $q(u_m)$}
We update the location of the $m\mbox{th}$ document using gradient
ascent, which takes the general form $\hu_m' = \hu_m +
\rho\nabla_{\hu_m}\mathcal{L}$. We take several steps in updating this
value within an iteration. For step $s$ we update $\hu_m$ as
\begin{equation}\label{eqn.upU}
 \hu_m^{(s+1)} = (1-\rho_s)\hu_m^{(s)} + \rho_s\sum_{k=1}^T\left(\mathbb{E}_Q[Z_k]\mbox{e}^{-\hell_k^T \hu_m^{(s)}} - \hbeta p_k\right)\hell_k.
\end{equation}
We let the step size $\rho$ be a function of step number $s$, and (for
example) set it to $\rho_s = \frac{1}{T} (3 + s)^{-1}$ for
$s=1,\dots,20$. We use $1/T$ to give a per-topic average, which helps
to stabilize the magnitude of the gradient by removing its dependence
on truncation level $T$, while $(3+s)^{-1}$ shrinks the step size. For
each iteration, we reset $s=1$.

\subsubsection*{\textbf{Corpus-level parameters}}

After optimizing the variational parameters for each document, we turn
to the corpus-level parameters. In the coordinate ascent algorithm, we
update each corpus-level parameter once before returning to the
document-level parameters.

\paragraph{Coordinate update of $q(\eta_k)$} The variational
distribution for the topic parameters is Dirichlet with parameter
vector $\gamma_k$. For each of $d=1,\dots,D$ vocabulary words
\begin{equation}\label{eqn.upEta}
  \gamma_{k,d} = \gamma_0  + \sum_{m=1}^M\sum_{n=1}^{N_m} \phi_{n,k}^{(m)}\mathbb{I}\(X_n^{(m)} =
  d\),
\end{equation}
where $\gamma_0$ is the parameter for the base distribution $\eta_k
\sim {\rm Dirichlet}(\gamma_0)$. Statistics needed for this term can
be updated in unison with updates to $q(C_n^{(m)})$ for faster
inference.

\paragraph{Coordinate update of $q(V_k)$} For $k=1,\dots,T-1$, the $q$
distribution for each $V_k$ is a delta function, $\delta_{\hV_k}$. The
truncation of the top-level DP results in $V_T := 1$. We use steepest
ascent to jointly optimize $\hV_1,\dots,\hV_{T-1}$. The gradient of
each element is
\begin{equation}\label{eqn.dqV}
  \frac{\partial \mathcal{L}(\cdot)}{\partial \hV_k} = -
  ~\frac{\halpha-1}{1-\hV_k} +
  \hbeta\[ \sum_m \( \E_Q[\ln
    Z_k^{(m)}] - \hell_k^T \hu_m\)  - M\psi(\hbeta p_k)\]\[\frac{p_k}{\hV_k}-\sum_{j > k} \frac{p_{j}}{1-\hV_k}\]     
\end{equation}
We observed similar performance using Newton's method in our experiments.

\paragraph{Coordinate update of $q(\ell_k)$}
We update the location of the $k\mbox{th}$ topic by gradient ascent,
which has the general form $\hell_k' = \hell_k +
\rho\nabla_{\hell_k}\mathcal{L}$. We use the same updating approach as
discussed for $\hu_m$. For step $s$ within a given iteration, the
update is
\begin{equation}\label{eqn.upL}
  \hell_k^{(s+1)} = (1-\rho_s/c)\hell_k + \rho_s\sum_{m=1}^M\left(\mathbb{E}_Q[Z_k]\mbox{e}^{-\hu_m^T\hell_k^{(s)}} - \hbeta p_k\right)\hu_m.
\end{equation}
As with $\hu_m$, we let the step size $\rho$ be a function of step
number $s$, and set it to $\rho_s = \frac{1}{M} (3 + s)^{-1}$.

\paragraph{Coordinate updates of $q(\alpha)$ and $q(\beta)$}
We place a $\mathrm{Gamma}(\tau_1,\tau_2)$ prior on $\alpha$ and model
the posterior with a delta function. The update for this parameter is
\begin{equation}\label{eqn.alpha_up}
\halpha  = \frac{K+\tau_1-2}{\tau_2 -\sum_{k=1}^{K-1}\ln(1-\hV_k)}
\end{equation}
In our empirical study we set $\tau_1 = 1$ and $\tau_2 = 10^{-3}$.

We also place a $\mathrm{Gamma}(\kappa_1,\kappa_2)$ prior on the second-level
concentration parameter $\beta$ and optimize using gradient ascent. The first
derivative is
\begin{equation}\label{eqn.upBeta}
\frac{\partial \mathcal{L}(\cdot)}{\partial \hbeta} = \sum_{m,k}p_k\(\psi(a_k^{(m)}) - \ln b_k^{(m)} - \ell_k^T u_m - \psi(\hbeta p_k)\) - \frac{\kappa_1-1}{\hbeta} - \kappa_2 .
\end{equation}
We set $\kappa_1 = 1$ and $\kappa_2 = 10^{-3}$.

\subsection{Stochastic variational inference}\label{sec.online}

The algorithm of Section \ref{sec:vbalg} can be called a
\textit{batch} algorithm because it updates all document-level
parameters in one ``batch'' before updating the global parameters. A
potential drawback of this batch inference approach for DILN
(as well as potential Monte Carlo sampling algorithms) is that the
per-iteration running time increases with an increasing number of
groups.  For many modeling applications, the algorithm may be
impractical for large-scale problems.

One solution to the large-scale data problem is to sub-sample a
manageable number of groups from the larger collection, and assume
that this provides a good statistical representation of the entire
data set. Indeed, this is the hope with batch inference, which views
the data set as a representative sample from the larger, unseen
population. However, in this scenario information contained in the
available data set may be lost. Stochastic variational inference
methods \citep{Hoffman:2010, Wang:2011, Sato:2001} aim for the best of
both worlds, allowing one to fit global parameters for massive
collections of data in less time than it takes to solve problems of
moderate size in the batch setting.

The idea behind stochastic variational inference is to perform
stochastic optimization of the variational objective function in
Equation (\ref{eqn.lowerbound}). In topic modeling, we can construe this
objective function as a sum over per-document terms and then obtain
noisy estimates of the gradients by evaluating them on sets of
documents sampled from the full corpus.  By following these noisy
estimates of the gradient with a decreasing step size, we are
guaranteed convergence to a local optimum of the variational objective
function \citep{Robbins:1951,Sato:2001,Hoffman:2010}.

Algorithmically, this gives an advantage over the optimization
algorithm of Section \ref{sec:vbalg} for large-scale machine learning.
The bottleneck of that algorithm is the variational ``E step,'' where
the document-level variational parameters are optimized for
\textit{all} documents using the current settings of the corpus-level
variational parameters (i.e., the topics and their locations, and
$\alpha$, $\beta$).  This computation may be wasteful, especially in
the first several iterations, where the initial topics likely do not represent the corpus well.  In contrast, the structure
of a stochastic variational inference algorithm is to repeatedly
subsample documents, analyze them, and then use them to update the
corpus-level variational parameters.  When the data set is massive,
these corpus-level parameters can converge before seeing any document
a second time.

In more detail, let $\bX$ be a very large collection of $M$
documents. We separate the hidden variables $\bTheta$ into those for
the top-level $\bTheta' =
\{\eta_{1:T},V_{1:T-1},\ell_{1:T},\alpha,\beta\}$ and the
document-level $\bTheta_m = \{C^{(m)}_{1:N_m},u_m,Z^{(m)}_{1:T}\}$ for
$m=1,\dots,M$. These variables have variational
parameters $\bPsi' =
\{\gamma_{1:T,1:D},\ell_{1:T},\hat{V}_{1:T-1},\hat{\alpha},\hat{\beta}\}$
and $\bPsi_m = \{\phi^{(m)}_{1:N_m},a^{(m)}_{1:T},b^{(b)}_{1:T},u_m\}$
for their respective $Q$ distributions. Because of the independence
assumption between documents, the variational objective decomposes
into a sum over documents,
\begin{equation}\label{eqn.online_batch}
 \mathcal{L}(\bX,\bPsi) = \sum_{m=1}^M \mathbb{E}_Q[\ln p(\bX_m,\bTheta_m,\bTheta')] + \sum_{m=1}^M \mathbb{H}[Q(\bTheta_m)] + \mathbb{H}[Q(\bTheta')].
\end{equation}
As we discussed, in batch inference we optimize variational
distributions on $\bTheta_1,\dots,\bTheta_M$ before updating those on
$\bTheta'$. Now, consider an alternate objective function at iteration
$t$ of inference,
\begin{equation}\label{eqn.online_onedoc}
 \mathcal{L}^{(t)}(\bX_{m_t},\bPsi_{m_t},\bPsi') = M\mathbb{E}_Q[\ln p(\bX_{m_t},\bTheta_{m_t}|\bTheta')] + M\mathbb{H}[Q(\bTheta_{m_t})] + \mathbb{E}_Q[\ln p(\bTheta')] + \mathbb{H}[Q(\bTheta')],
\end{equation}
where $m_t$ is selected uniformly at random from $\{1,\dots,M\}$. An
approach to optimize this objective function would be to first
optimize the variational parameters of $Q(\bTheta_{m_t})$, followed by
a single gradient step for those of $Q(\bTheta')$. In determining the
relationship between Equation (\ref{eqn.online_onedoc}) and Equation
(\ref{eqn.online_batch}), note that under the uniform distribution
$p(m_t)$ on which document is selected,
\begin{equation}
 \mathbb{E}_{p(m_t)}[\mathcal{L}^{(t)}(\bX_{m_t},\bPsi_{m_t},\bPsi')] = \mathcal{L}(\bX,\bPsi).
\end{equation}
We are thus \emph{stochastically} optimizing $\mathcal{L}$. In
practice, one document is not enough to ensure fast convergence of
$Q(\bTheta')$. Rather, we select a subset $B_t \subset \{1,\dots,M\}$
at iteration $t$ and optimize
\begin{eqnarray}\label{eqn.online_subset}
 \mathcal{L}^{(t)}(\bX_{B_t},\bPsi_{B_t},\bPsi') &=& \frac{M}{|B_t|}\sum_{i\in B_t}\mathbb{E}_Q[\ln p(\bX_i,\bTheta_i|\bTheta')] + \frac{M}{|B_t|} \sum_{i\in B_t}\mathbb{H}[Q(\bTheta_i)] \nonumber\\
&& + ~\mathbb{E}_Q[\ln p(\bTheta')] ~+~ \mathbb{H}[Q(\bTheta')],
\end{eqnarray}
over the variational parameters of $Q(\bTheta_{B_t})$. We again follow
this with a step for the variational parameters of $Q(\bTheta')$, but
this time using the information from documents indexed by $B_t$. That
is, for some corpus-level parameter $\psi \in \bPsi'$, the update of
$\psi$ at iteration $t+1$ given $\psi$ at iteration $t$ is
\begin{equation}\label{eqn.generic_grad}
 \psi^{(t+1)} = \psi^{(t)} + \rho_t\bA_{\psi}\nabla_{\psi} \mathcal{L}^{(t)}(\bX_{B_t},\bPsi_{B_t}, \bPsi') ,
\end{equation}
where $\bA_{\psi}$ is a positive definite preconditioning matrix and $\rho_t > 0$ is a step size satisfying
\begin{equation}
 \sum_{t=1}^{\infty} \rho_t = \infty, \quad\quad \sum_{t=1}^{\infty} \rho_t^2 < \infty .
\end{equation}
In our experiments, we select the form $\rho_t =
(\zeta + t)^{-\kappa}$ with $\kappa \in (0.5,1]$ and $\zeta > 0$.

In some cases, the preconditioner $\bA_{\psi}$ can be set to give
simple and clear updates. For example, in the case of topic modeling,
\cite{Hoffman:2010} show how the inverse Fisher information leads to
very intuitive updates (see the next section). This is a special case
of the theory outlined in \cite{Sato:2001} that arises in conjugate
exponential family models. However, the Fisher
information is not required for stochastic variational inference; we can precondition with the inverse negative Hessian or
decide not to precondition.

\begin{algorithm}[tb]
   \caption{Stochastic variational Bayes for DILN}
   \label{alg:alg2}
\raggedright
   Stochastically optimize the variational lower bound $\mathcal{L}$\\
   Primary goal: Optimize corpus-wide variational parameters $\bPsi'$\\
   Secondary goal: Optimize document-specific parameters $\bPsi_m$ for $m=1,\dots,M$\\
\begin{algorithmic}[1]
  \WHILE{$\bPsi'$ has not converged}
  \STATE Select random subset $B_t\subset\{1,\dots,M\}$
  \FOR{$m \in B_t$}
  \STATE Optimize $\bPsi_m$ ({Equations \ref{eqn.upC}--\ref{eqn.upU}})
  \ENDFOR
  \STATE Set gradient step size $\rho_t = (\zeta + t)^{-\kappa}$, $\kappa \in (\frac{1}{2},1]$
  \STATE Update $\bPsi'$ using gradient of $\mathcal{L}^{(t)}$ constructed from documents $m\in B_t$ (Equations \ref{eqn.dqV}, \ref{eqn.upL}, \ref{eqn.upBeta}, \ref{eqn.generic_grad}, \ref{eqn.stochtopicup}--\ref{eqn.dqVHess})
  \ENDWHILE
  \STATE Optimize $\bPsi_m$ for $m=1,\dots,M$ using optimized $\bPsi'$
\end{algorithmic}
\end{algorithm}

\subsubsection{The stochastic variational inference algorithm for
  DILN}

The stochastic algorithm selects a subset of documents at step $t$,
coded by a set of index values $B_t$, and optimizes the document-level
parameters for these documents while holding all corpus-level
parameters fixed.  These parameters are the word indicators $C_k^{(m)}$, the unnormalized topic weights $Z_k^{(m)}$ and the document locations $u_k$. (See Section \ref{sec:vbalg} for discussion on inference for these variables.) Given the values of the document-level variational parameters for documents indexed by $B_t$, we
now describe the corpus-level updates in the stochastic inference
algorithm. Algorithm \ref{alg:alg2} summarizes this general inference structure.

\paragraph{Stochastic update of $q(\eta_k)$} This update follows from
\cite{Hoffman:2010} and \cite{Wang:2011}. We set $\bA_{\gamma_k}$ to
be the inverse Fisher information of $q(\eta_k)$,
\begin{equation*}
  \bA_{\gamma_k} =
  \left(-\frac{\partial^2 \ln
      q(\eta_k)}{\partial\gamma_k\partial\gamma_k^T}\right)^{-1}.
\end{equation*}
With this quantity, we take the product
$\bA_{\gamma_k}\nabla_{\gamma_k}\mathcal{L}^{(t)}(\bX_{B_t},\bPsi_{B_t},\bPsi')$.
This leads to give the following update for each $\gamma_{k,d}$,
\begin{equation}\label{eqn.stochtopicup}
  \gamma_{k,d}^{(t+1)} = (1-\rho_t)\gamma_{k,d}^{(t)} + \rho_t\left(\gamma_0 + \frac{M}{|B_t|} \sum_{n,m\in B_t} \phi_{n,k}^{(m)}\I(X_n^{(m)} = d)\right).
\end{equation}
In this case, premultiplying the gradient by the inverse Fisher
information cancels the Fisher information in the gradient and thus
removes the cross-dependencies between the components of $\gamma_k$.
We use preconditioning to simplify the computation, rather than to
speed up optimization.  See~\cite{Hoffman:2010}, \cite{Wang:2011} and
\cite{Sato:2001} for details.

\paragraph{Stochastic update of $q(V_k)$ and $q(\ell_k)$} The
stochastic updates of the delta $q$ distributions do not use the
Fisher information. Rather, we update the vectors $V =
[V_1,\dots,V_{T-1}]^T$ and $\ell_k$ for $k = 1,\dots,T$ by taking
steps in their Newton directions using the data in batch $B_t$ to
determine this direction. The gradients $\nabla\mathcal{L}$ for these
parameters are given in the batch algorithm and their form is
unchanged here. The key difference is that the gradient of these
parameters at step $t$ is only calculated over documents with index
values in $B_t$. We use the inverse negative Hessian as a preconditioning matrix for $\hell_k$ and $(\hV_1,\dots,\hV_{T-1})$. For $\ell_k$, the preconditioning matrix is
\begin{equation}
  \bA_{\hell_k}^{-1} =  c^{-1}I + \sum_{m=1}^M(\E_Q[Z_k]\mbox{e}^{-\hell_k \hu_m})\hu_m \hu_m^T .
\end{equation}
For $(\hV_1,\dots,\hV_{T-1})$ the values of $(\bA_{\hV}^{-1})_{kk}$ and $(\bA_{\hV}^{-1})_{kr}$ are found from the second derivatives (with the second derivatives written for $r < k$)
\begin{eqnarray}\label{eqn.dqVHess}
  -\frac{\partial^2 \mathcal{L}(\cdot)}{\partial \hV_k^2} &=& \frac{\halpha-1}{(1-\hV_k)^2} +  \hbeta^2 M\psi'(\hbeta p_k)\frac{p_k}{\hV_k}\(\frac{p_k}{\hV_k}-\sum_{j > k} \frac{p_{j}}{1-\hV_k}\), \\
  -\frac{\partial^2 \mathcal{L}(\cdot)}{\partial \hV_k\partial \hV_r} &=&-~\hbeta^2M\psi'(\hbeta p_k)\frac{p_k}{(1-\hV_r)}\(\frac{p_k}{\hV_k}-\sum_{j > k} \frac{p_{j}}{1-\hV_k}\)~+\hspace{1.25in}
\end{eqnarray}
\begin{equation}\nonumber
\hbeta\[ \sum_m \( \E_Q[\ln Z_k^{(m)}] - \hell_k^T \hu_m\)  - M\psi(\hbeta p_k)\]\[\frac{p_k}{\hV_k(1-\hV_r)}-\sum_{j > k} \frac{p_{j}}{(1-\hV_k)(1-\hV_r)}\].
\end{equation}

\paragraph{Online update of $q(\alpha)$ and $q(\beta)$} The stochastic
updates for $\hat{\beta}$ move in the direction of steepest ascent,
calculated using the documents in the batch. Since this is a
one-dimensional parameter, we optimize a batch-specific value for this
parameter at step $t$, $\tilde{\beta}_t$, and set $\hat{\beta}_{t+1} =
(1-\rho_t)\hat{\beta}_t + \rho_t \tilde{\beta}_t$. The update for $\hat{\alpha}$
does not consider document-level parameters, and so this value follows
the update given in Equation (\ref{eqn.alpha_up}).

\subsection{A new variational inference algorithm for the
  HDP}\label{sec:dilnvshdp}

The variational inference algorithm above relates closely to one that
can be derived for the HDP using the normalized gamma process
representation of Section \ref{sec:HDP}. The difference lies in the
update for the topic weight $q(Z_k^{(m)})$ in Equation (\ref{eqn.upZ}).  In both
algorithms, the update for its variational parameter $a_k^{(m)}$ contains the prior from the
top-level DP, and the expected number of words in document $m$ drawn
from topic $k$. The variational parameter $b^{(m)}$ distinguishes DILN from the
HDP.

We can obtain a variational inference algorithm for the HDP by setting the
first term in the update for $b_k^{(m)}$ equal to one. In contrast,
the first term for DILN is $\exp\{-\hell_k^T\hu_m\}$, which is the
Gaussian process that generates the covariance between component
probability weights.  Including or excluding this term switches
between variational inference for DILN and variational inference for
the HDP. See the appendix for a fuller derivation.

\subsection{MCMC inference}

Markov chain Monte Carlo~\citep[MCMC,][]{Robert:2004} sampling is a
more common strategy for approximate posterior inference in Bayesian
nonparametric models, and for the hierarchical Dirichlet process in
particular.  In MCMC methods, samples are drawn from a carefully
designed Markov chain, whose stationary distribution is the target
posterior of the model parameters.  MCMC is convenient for the many
Bayesian nonparametric models that are amenable to Gibbs sampling,
where the Markov chain iteratively samples from the conditional
distribution of each latent variable given all of the other latent
variables and the observations.

However, Gibbs sampling is not an option for DILN because the Gaussian
process component does not have a closed-form full conditional
distribution.  One possible sampling algorithm for DILN inference
would use Metropolis-Hastings~\citep{Hastings:1970}, where samples are
drawn from a proposal distribution and then accepted or rejected.
Designing a good proposal distribution is the main problem in
designing Metropolis-Hastings algorithms, and in DILN this problem is
more difficult than usual because the hidden variables are highly
correlated.

Recently, slice sampling has been applied to sampling of infinite
mixture models by turning the problem into a finite sampling problem
\citep{Griffin:2010,Maria:2011}.  These methods apply when the mixture
weights are either from a simple stick-breaking prior or a normalized
random measures that can be simulated from a Poisson process.  Neither
of these settings applies to DILN because the second-level DP is a
product of a DP and an exponentiated GP.  Furthermore, it is not clear
how to extend slice sampling methods to hierarchical models like the
HDP or DILN.

Variational methods mitigate all these issues by using optimization to
approximate the posterior.  Our algorithm sacrifices the theoretical
(and eventual) convergence to the full posterior in favor of a simpler
distribution that is fit to minimize its KL-divergence to the
posterior.  Though we must address issues of local minima in the
objective, we do not need to develop complicated proposal
distributions or solve the difficult problem of assessing convergence
of a high-dimensional Markov chain to its stationary
distribution.\footnote{Note our evaluation method of
  Section~\ref{sec.results} does not use the divergence of the
  variational approximation and the true posterior. Rather, we measure
  the corresponding approximation to the predictive distribution.  On
  a pilot study of batch inference, we found that MCMC inference (with
  its approximate predictive distribution) did not produce
  distinguishable results from variational inference.} Furthermore,
variational inference is ideally suited to the stochastic optimization
setting, allowing for approximate inference with very large data sets.

\section{Empirical study} \label{sec.results} We evaluate the DILN
topic model with both batch and stochastic inference. For batch
inference, we compare with the HDP and correlated topic model (CTM) on
four text corpora: \textit{The Huffington Post}, \textit{The New York
  Times}, \textit{Science} and \textit{Wikipedia}. We divide each
corpus into five training and testing groups selected from a larger
set of documents (see Table \ref{tab.corpus}).

For stochastic inference, we use the \textit{Nature} corpus to assess
performance. This corpus contains 352,549 documents spanning
1869-2003; we used a vocabulary of 4,253 words. We compare stochastic
DILN with a stochastic HDP algorithm and with online
LDA~\citep{Hoffman:2010}.

\begin{table}[t]
\caption{Data sets. Five training/testing sets were
constructed by selecting the number of documents shown for each corpus from
larger data sets.\vspace{1mm}}\label{tab.corpus}
\centering
\begin{tabular}{lllll} \hline Corpus & \# training & \# testing & vocabulary size & \# total words \\ \hline Huffington Post & ~~3,000 & ~~1,000 & ~~6,313 & ~~660,000 \\ New York Times & ~~5,000 & ~~2,000 & ~~3,012 & ~~720,000 \\ Science & ~~5,000 & ~~2,000 & ~~4,403 & ~~1,380,000 \\ Wikipedia & ~~5,000 & ~~2,000 & ~~6,131 & ~~1,770,000 \\ \hline
\end{tabular}
\end{table}

\subsection{Evaluation metric}\label{sec.evaluation}

Before discussing the experimental setup and results, we discuss our method for
evaluating performance. We evaluate the approximate posterior of all models by
measuring its predictive ability on held-out documents.  Following
\cite{Asuncion:2009}, we randomly partition each test document into two halves
and evaluate the conditional distribution of the second half given the first
half and the training data.  Operationally, we use the first half of each
document to find estimates of document-specific topic proportions and then
evaluate how well these combine with the fitted topics to predict the second
half of the document.

More formally, denote the training data by ${\cal D}$, a test document
as $\bX$, which is divided into halves $\bX'$ and $\bX''$.  We
want to calculate the conditional marginal probability,
\begin{equation}
  \label{eq:test}
  p({\bX}''| {\bX}', {\cal D}) =
  \int_{\Omega_{\boldsymbol{\eta},\boldsymbol{Z}}}\prod_{n=1}^{N}\left\lbrace\sum_{k=1}^T
    p({X}''_n|\eta_k)p({C}''_n = k |{Z_{1:T}})\right\rbrace
  dQ(\boldsymbol{Z})dQ(\boldsymbol{\eta})
\end{equation}
where $N$ is the number of observations constituting $\bX''$, $C_n''$
is the latent indicator associated with the $n$th word in $\bX''$, and
$\boldsymbol{\eta} := \eta_{1:T}$ and $\boldsymbol{Z} := Z_{1:T}$.

Since the integral in Equation (\ref{eq:test}) is intractable, we sample
i.i.d.\ values from the factorized distributions $Q({Z_{1:T}})$ and
$Q(\eta_{1:T})$ for approximation. We note that the information
regarding the document's correlation structure can be found in $Q({Z_{1:T}})$.

We then use this approximation of the marginal likelihood to compute
the average per-word perplexity for the second half of the test
document,
\begin{equation}
{\rm perplexity} = \exp\left\lbrace \frac{-\ln p({\bX}''| {\bX}')}{N} \right\rbrace ,
\end{equation}
with lower perplexity indicating better performance. Note that the
term $\ln p({\bX}''| {\bX}')$ involves a sum over the $N$ words in
$\bX''$.  Also note that this is an objective measure of the
predictive performance of the predictive probability distribution
computed from the variational approximation.  It is a good measure of
performance (of the model and the variational inference algorithm)
because it does not rely on the closeness of the variational
distribution to the true posterior, as measured by the variational lower bound. That closeness, much like whether
a Markov chain has converged to its stationary distribution, is
difficult to assess.

\begin{figure}[t!]
\centering
\includegraphics[width=1\textwidth]{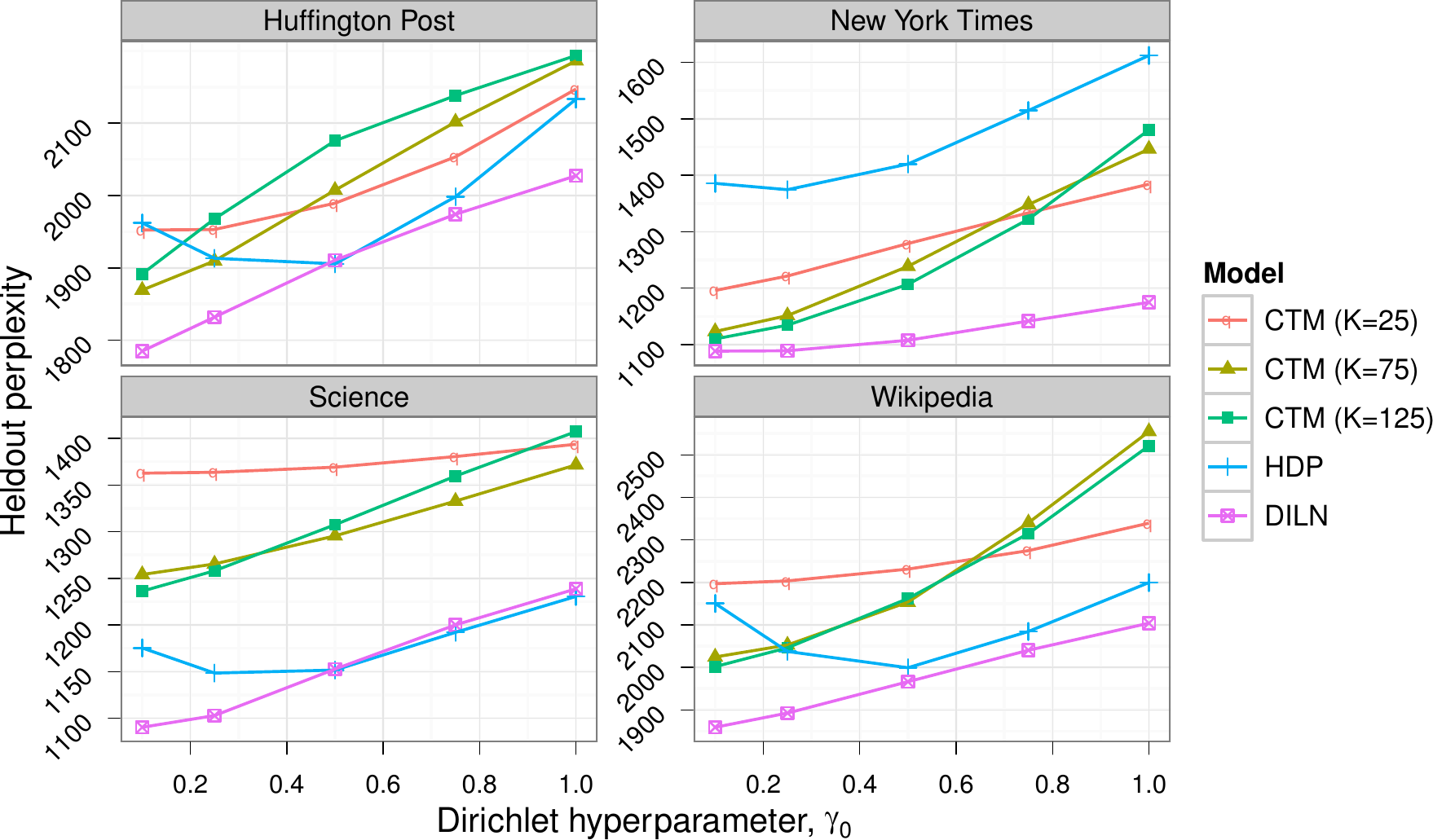}
\caption{Perplexity results for four text corpora and averaged over
  five training/testing sets.  For a fixed Dirichlet hyperparameter,
  the DILN topic model typically achieves better perplexity than both
  the HDP and CTM models.  In all corpora, DILN achieves the best perplexity
  overall. }\label{fig.experiments}
\end{figure}

\subsection{Experimental setup and results}

\paragraph{Batch variational inference experiments} We trained all
models using variational inference; for the CTM, this is the algorithm
given in \cite{Blei:2007}; for the HDP, we use the inference method
from \mysec{vb}.  For DILN, we use a latent space with $d = 20$
and set the location variance parameter $c = 1/20$. For DILN and the
HDP, we truncate the top-level stick-breaking construction at $T=200$
components. For the CTM, we consider $K \in \{20, 50, 150\}$ topics.
In our experiments, both DILN and HDP used significantly fewer topics
than the truncation level, indicating that the truncation level was
set high enough.  The CTM is not sparse in this sense.

We initialize all models in the same way; to initialize the
variational parameters of the topic Dirichlet, we first cluster the
empirical word distributions of each document with three iterations of
k-means using the $L_1$ distance measure. We then reorder these topics
by their usage according to the indicators produced by k-means. We
scale these k-means centroids and add a small constant plus noise to
smooth the initialization. The other parameters are initialized to
values that favor a uniform distribution on these topics.  Variational
inference is terminated when the fractional change in the lower bound
of Equation (\ref{eqn.lowerbound}) falls below $10^{-3}$. We run each
algorithm using five different topic Dirichlet hyperparameter
settings: $\gamma_0 \in \{0.1, 0.25, 0.5, 0.75, 1.0\}$.

Figure \ref{fig.experiments} contains testing results for the four
corpora. In general, DILN outperforms both the HDP and CTM. Given that
the inference algorithms for DILN and the HDP are only different in the
one term discussed in Section \ref{sec:dilnvshdp}, this demonstrates
that the latent location space models a correlation structure that helps in predicting words.  Computation time
for DILN and the HDP was comparable, both requiring on the order of
one minute per iteration. Depending on the truncation level, the CTM
was slightly to significantly faster than both DILN and the HDP.

\begin{figure}[h!]\vspace{1cm}
\begin{tabular}{cc}
 \includegraphics[width=.465\textwidth]{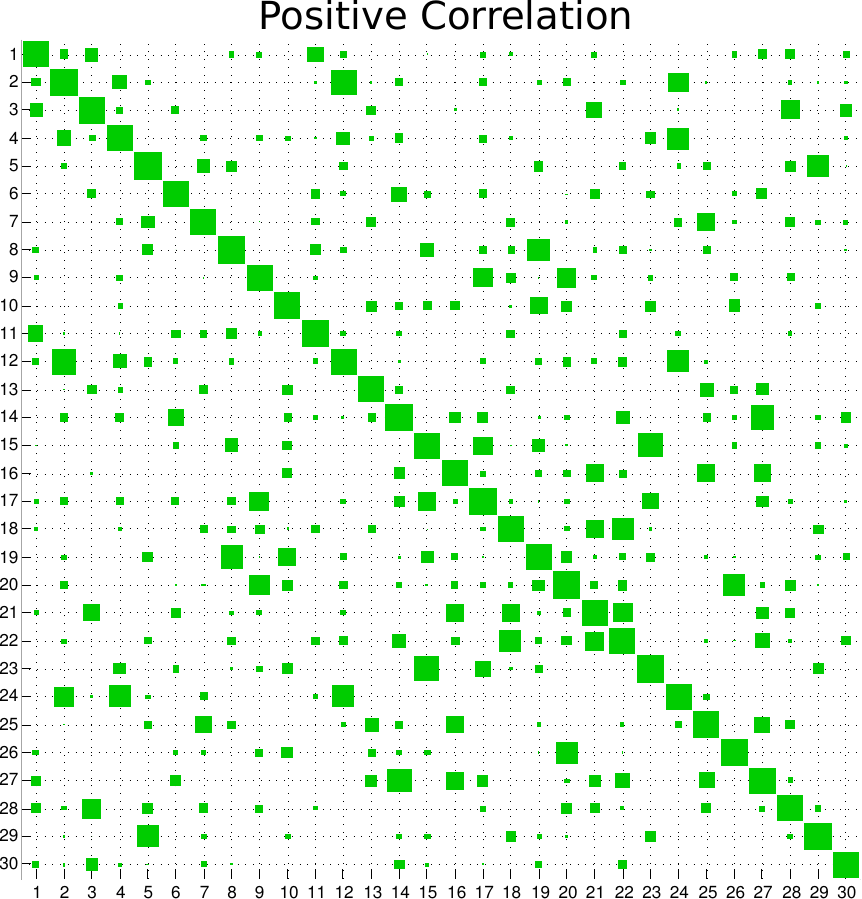}& \includegraphics[width=.465\textwidth]{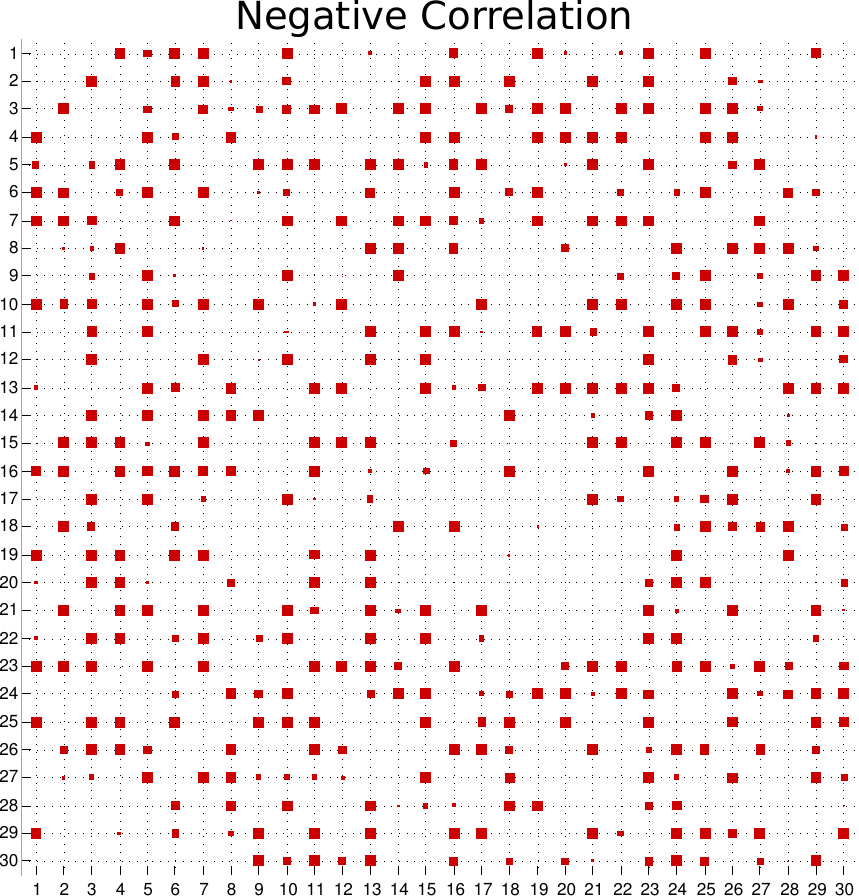}\\
\end{tabular}\vspace{2mm}
\begin{scriptsize}
\begin{tabular}{l}\hline\vspace{-2mm}\\ 
Topic 1: campaign, democratic, candidate, republican, election, voter, political, presidential, vote, party\\
Topic 2: game, victory, second, score, third, win, team, play, season, lose\\
Topic 3: president, executive, chief, vice, name, director, advertising, chairman, senior, company\\
Topic 4: team, player, season, coach, game, play, football, league, contract, sign\\
Topic 5: add, heat, pound, cup, oil, minute, water, large, dry, serve\\
Topic 6: building, build, house, space, site, project, construction, area, foot, plan\\
Topic 7: drug, patient, treatment, study, disease, risk, health, treat, cancer, cause\\
Topic 8: economy, economic, percent, growth, increase, government, states, economist, price, rate\\
Topic 9: police, officer, arrest, man, charge, yesterday, official, crime, drug, release\\
Topic 10: share, company, stock, buy, percent, investment, acquire, sell, investor, firm\\
Topic 11: budget, tax, cut, increase, taxis, state, plan, propose, reduce, pay\\
Topic 12: shot, point, play, game, hit, ball, night, shoot, player, put\\
Topic 13: computer, internet, information, site, technology, system, software, online, user, program\\
Topic 14: art, artist, museum, exhibition, painting, collection, gallery, design, display, sculpture\\
Topic 15: government, political, country, international, leader, soviet, minister, states, foreign, state\\
Topic 16: book, story, write, novel, author, life, woman, writer, storey, character\\
Topic 17: attack, kill, soldier, bomb, bombing, area, official, report, group, southern\\
Topic 18: song, sing, band, pop, rock, audience, singer, voice, record, album\\
Topic 19: market, stock, price, fall, trading, dollar, investor, trade, rise, index\\
Topic 20: trial, lawyer, charge, prosecutor, case, jury, guilty, prison, sentence, judge\\
Topic 21: play, movie, film, star, actor, character, theater, role, cast, production\\
Topic 22: dance, stage, perform, dancer, company, production, present, costume, theater, performance\\
Topic 23: peace, israeli, palestinian, talk, palestinians, territory, arab, leader, visit, settlement\\
Topic 24: guy, thing, lot, play, feel, kind, game, really, little, catch\\
Topic 25: science, theory, scientific, research, human, suggest, evidence, fact, point, question\\
Topic 26: court, law, state, legal, judge, rule, case, decision, appeal, lawyer\\
Topic 27: image, photograph, picture, view, photographer, subject, figure, paint, portrait, scene\\
Topic 28: report, official, member, commission, committee, staff, agency, panel, investigate, release\\
Topic 29: wine, restaurant, food, menu, price, dish, serve, meal, chicken, dining\\
Topic 30: graduate, marry, father, degree, receive, ceremony, wedding, daughter, son, president\\\hline
\end{tabular}
\end{scriptsize}
\caption{\textit{New York Times}: The ten most probable words from the 30 most popular topics. At top are the positive and negative correlation coefficients for these topics calculated by taking the dot product of the topic locations, $\ell_k^T\ell_{k'}$ (separated for clarity).}\label{fig.nyt}
\end{figure}

\begin{figure}[h!]\vspace{1cm}
\begin{tabular}{cc}
 \includegraphics[width=.465\textwidth]{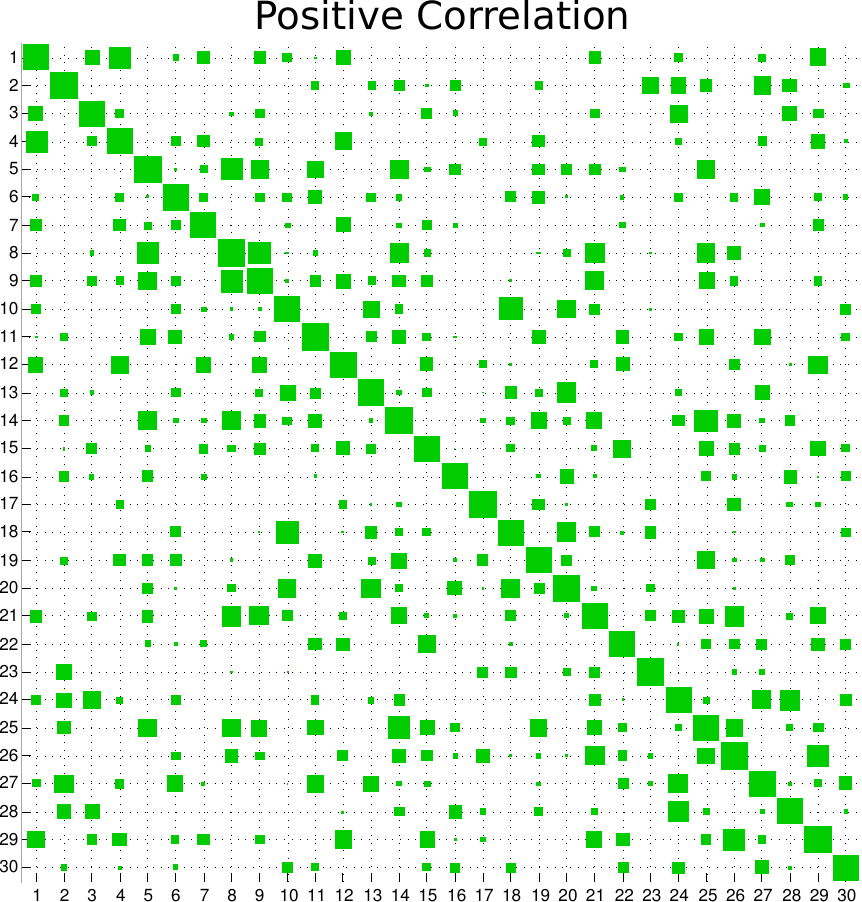}& \includegraphics[width=.465\textwidth]{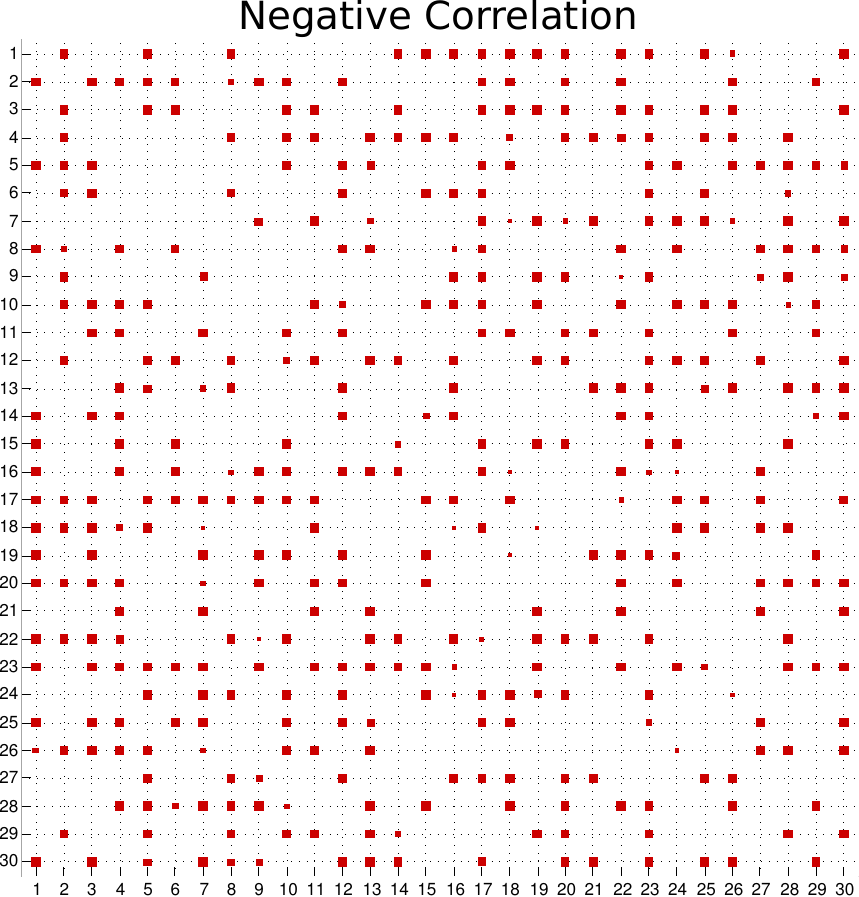}\\
\end{tabular}\vspace{2mm}
\begin{scriptsize}
\begin{tabular}{l}\hline\vspace{-2mm}\\ 
Topic 1: get, really, like, just, know, hes, think, dont, thing, say\\
Topic 2: percent, year, said, last, prices, economy, quarter, home, economic, housing\\
Topic 3: day, mother, life, family, father, mothers, love, time, home, fathers\\
Topic 4: make, like, dont, youre, people, time, get, see, love, just\\
Topic 5: delegates, obama, superdelegates, democratic, party, states, convention, primaries, michigan\\
Topic 6: mccain, john, mccains, republican, campaign, bush, hes, just, senator, said\\
Topic 7: show, song, said, music, night, first, david, like, simon, performance\\
Topic 8: clinton, obama, clintons, hillary, nomination, democratic, barack, race, obamas, supporters\\
Topic 9: hillary, obama, president, candidate, shes, win, time, democratic, hillarys, running\\
Topic 10: iran, nuclear, weapons, states, said, united, attack, bush, president, iranian\\
Topic 11: democrats, republican, republicans, election, democratic, house, vote, states, gop, political\\
Topic 12: words, word, people, power, like, language, point, written, person, powerful\\
Topic 13: iraq, war, american, bush, afghanistan, years, petraeus, troops, new, mission\\
Topic 14: voters, obama, indiana, carolina, north, clinton, polls, primary, democratic, pennsylvania\\
Topic 15: america, american, nation, country, americans, history, civil, years, king, national\\
Topic 16: said, city, people, two, homes, area, water, river, state, officials\\
Topic 17: media, news, story, coverage, television, new, public, journalism, broadcast, channel\\
Topic 18: israel, peace, israeli, east, hamas, palestinian, state, arab, middle, israels\\
Topic 19: poll, chance, gallup, degrees, winning, results, tracking, general, election, august\\
Topic 20: said, iraqi, government, forces, baghdad, city, shiite, security, sadr, minister\\
Topic 21: senator, obama, obamas, people, clinton, pennsylvania, comments, bitter, remarks, negative\\
Topic 22: rights, law, court, justice, constitution, supreme, right, laws, courts, constitutional\\
Topic 23: company, said, billion, yahoo, stock, share, inc, deal, microsoft, shares\\
Topic 24: health, care, families, insurance, working, pay, help, americans, plan, people\\
Topic 25: white, race, voters, obama, virginia, west, percent, states, whites, win\\
Topic 26: wright, obama, rev, jeremiah, pastor, obamas, reverend, political, said, black\\
Topic 27: tax, government, economic, spending, taxes, cuts, economy, budget, federal, people\\
Topic 28: study, cancer, found, drugs, age, risk, drug, heart, brain, medical\\
Topic 29: people, man, black, america, didnt, god, hope, know, years, country\\
Topic 30: global, climate, warming, change, energy, countries, new, carbon, environmental, emissions\\\hline
\end{tabular}
\end{scriptsize}
\caption{\textit{Huffington Post}: The ten most probable words from the 30 most popular topics. At top are the positive and negative correlation coefficients for these topics calculated by taking the dot product of the topic locations, $\ell_k^T\ell_{k'}$ (separated for clarity).}\label{fig.huff}
\end{figure}

\begin{figure}[h!]\vspace{1cm}
\begin{tabular}{cc}
 \includegraphics[width=.465\textwidth]{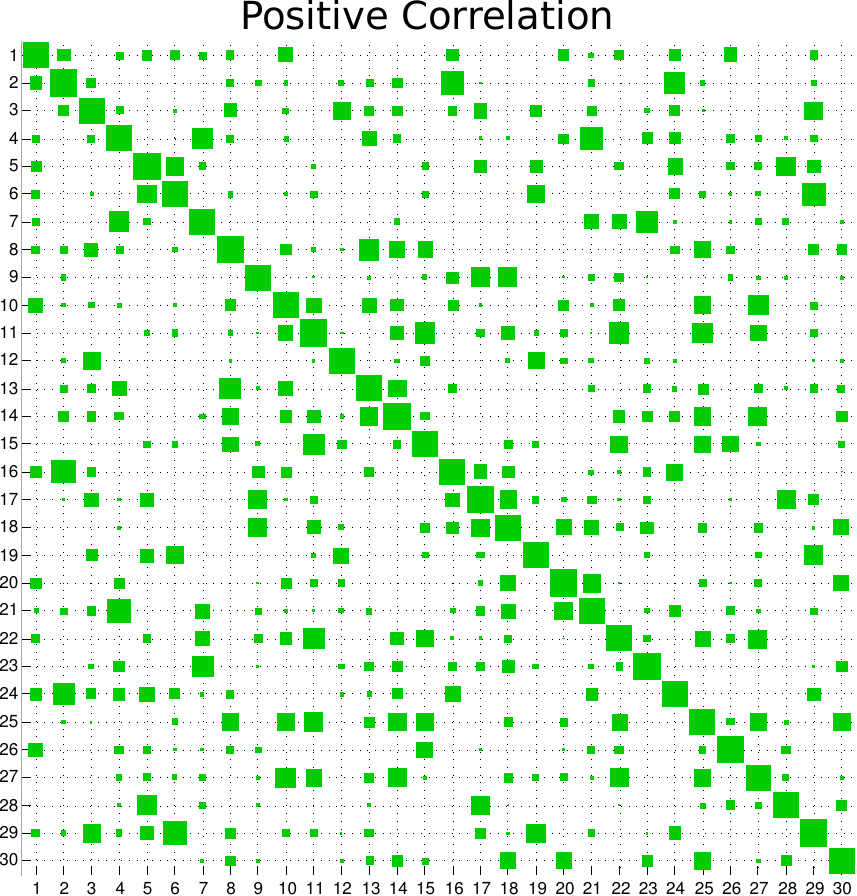}& \includegraphics[width=.465\textwidth]{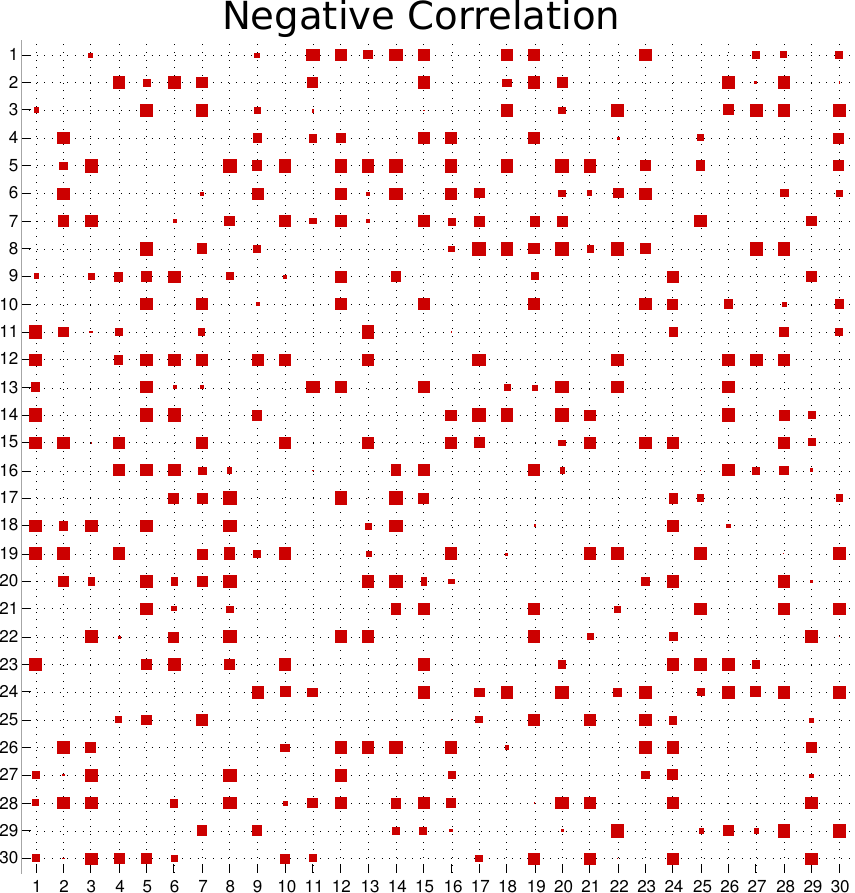}\\
\end{tabular}\vspace{2mm}
\begin{scriptsize}
\begin{tabular}{l}\hline\vspace{-2mm}\\ 
Topic 1: manager, science, fax, advertising, aaas, sales, recruitment, member, associate, washington\\
Topic 2: research, science, funding, scientists, university, universities, government, program, year\\ 
Topic 3: fault, plate, earthquake, earthquakes, zone, crust, seismic, fig, crustal, large\\
Topic 4: hiv, virus, infection, infected, viral, viruses, human, immunodeficiency, aids, disease\\
Topic 5: species, forest, forests, conservation, ecosystems, fish, natural, land, tropical, ecological\\
Topic 6: climate, changes, temperature, change, global, atmospheric, carbon, years, year, variability\\
Topic 7: cells, immune, cell, antigen, response, responses, mice, lymphocytes, antibody, specific\\
Topic 8: transcription, binding, dna, transcriptional, promoter, polymerase, factors, site, protein\\ 
Topic 9: says, university, just, colleagues, team, like, researchers, meeting, new, end\\
Topic 10: structure, residues, helix, binding, two, fig, helices, side, three, helical\\
Topic 11: proteins, protein, membrane, ras, gtp, binding, bound, transport, guanosine, membranes\\
Topic 12: pressure, temperature, high, phase, pressures, temperatures, experiments, gpa, melting\\ 
Topic 13: rna, mrna, site, splicing, rnas, pre, intron, base, cleavage, nucleotides\\
Topic 14: protein, cdna, fig, sequence, lane, purified, human, lanes, clone, gel\\
Topic 15: kinase, protein, phosphorylation, kinases, activity, activated, signaling, camp, pathway\\ 
Topic 16: university, students, says, faculty, graduate, women, science, professor, job, lab\\
Topic 17: new, says, university, years, human, humans, ago, found, modern, first\\
Topic 18: researchers, found, called, says, team, work, colleagues, new, university, protein\\
Topic 19: isotopic, carbon, oxygen, isotope, water, values, ratios, organic, samples, composition\\
Topic 20: disease, patients, diseases, gene, alzheimers, cause, mutations, syndrome, protein, genetic\\
Topic 21: aids, vaccine, new, researchers, vaccines, trials, people, research, clinical, patients\\
Topic 22: receptor, receptors, binding, ligand, transmembrane, surface, signal, hormone, extracellular\\ 
Topic 23: cells, cell, bone, human, marrow, stem, types, line, lines, normal\\
Topic 24: united, states, countries, international, world, development, japan, european, nations, europe\\
Topic 25: proteins, protein, yeast, two, domain, sequence, conserved, function, amino, family\\
Topic 26: letters, mail, web, end, new, org, usa, science, full, letter\\
Topic 27: amino, acid, peptide, acids, peptides, residues, sequence, binding, sequences, residue\\
Topic 28: species, evolution, evolutionary, phylogenetic, biology, organisms, history, different, evolved\\ 
Topic 29: ocean, sea, pacific, water, atlantic, marine, deep, surface, north, waters\\
Topic 30: gene, genes, development, genetic, mouse, function, expressed, expression, molecular, product\\\hline
\end{tabular}
\end{scriptsize}
\caption{\textit{Science}: The ten most probable words from the 30 most popular topics. At top are the positive and negative correlation coefficients for these topics calculated by taking the dot product of the topic locations, $\ell_k^T\ell_{k'}$ (separated for clarity).}\label{fig.sci}
\end{figure}

We display the learned correlation structure for the four corpora in
Figures \ref{fig.nyt}--\ref{fig.sci}. (see Figure \ref{fig.wiki} for results on a slightly larger \textit{Wikipedia} corpus.) In these
figures, we represent the 30 most probable topics by their ten most
probable words. Above these lists, we show the positive and negative
correlations learned using the latent locations $\ell_k$. For two
topics $i$ and $j$ this value is
$\ell_i^T\ell_{j}/\|\ell_i\|_2\|\ell_{j}\|_2$. From these figures, we
see that DILN learns meaningful underlying correlations in topic expression within a document.

\begin{figure}[h!]\vspace{1cm}
{\includegraphics[width=1\textwidth]{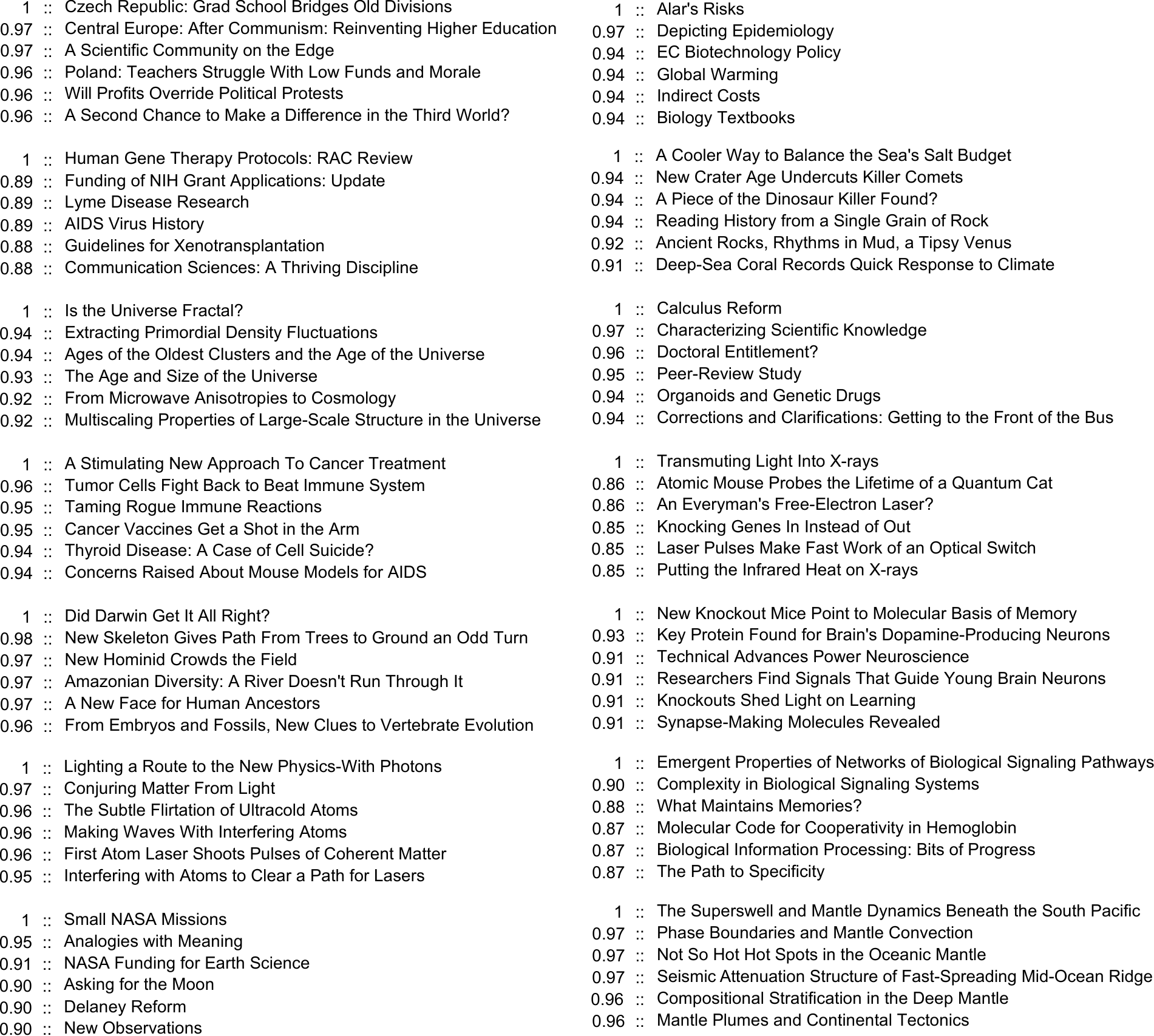}}
\caption{Several example document searches for \textit{Science}. The first document is the query document, followed by the most similar documents according to the cosine similarity measure on their locations (given at left).}\label{fig.sci_recs} 
\end{figure}
As we discussed in Section \ref{sec:kernel}, the underlying vectors
$u_m \in \realline^d$ associated with each document can be used for
retrieval applications. In Figure \ref{fig.sci_recs}, we show
recommendation lists for a 16,000 document corpus of the journal
\textit{Science} obtained using these underlying document
locations. We use the cosine similarity between two documents for
ranking, which for documents $i$ and $j$ is equal to
$u_i^Tu_j/\|u_i\|_2\|u_j\|_2$.  We show several lists of recommended
articles based on randomly selected query articles. These lists show
that, as with the underlying correlations learned between the topics,
DILN learns a meaningful relationship between the documents as well,
which is useful for navigating text corpora.

\begin{figure}[t]
{\includegraphics[width=1\textwidth]{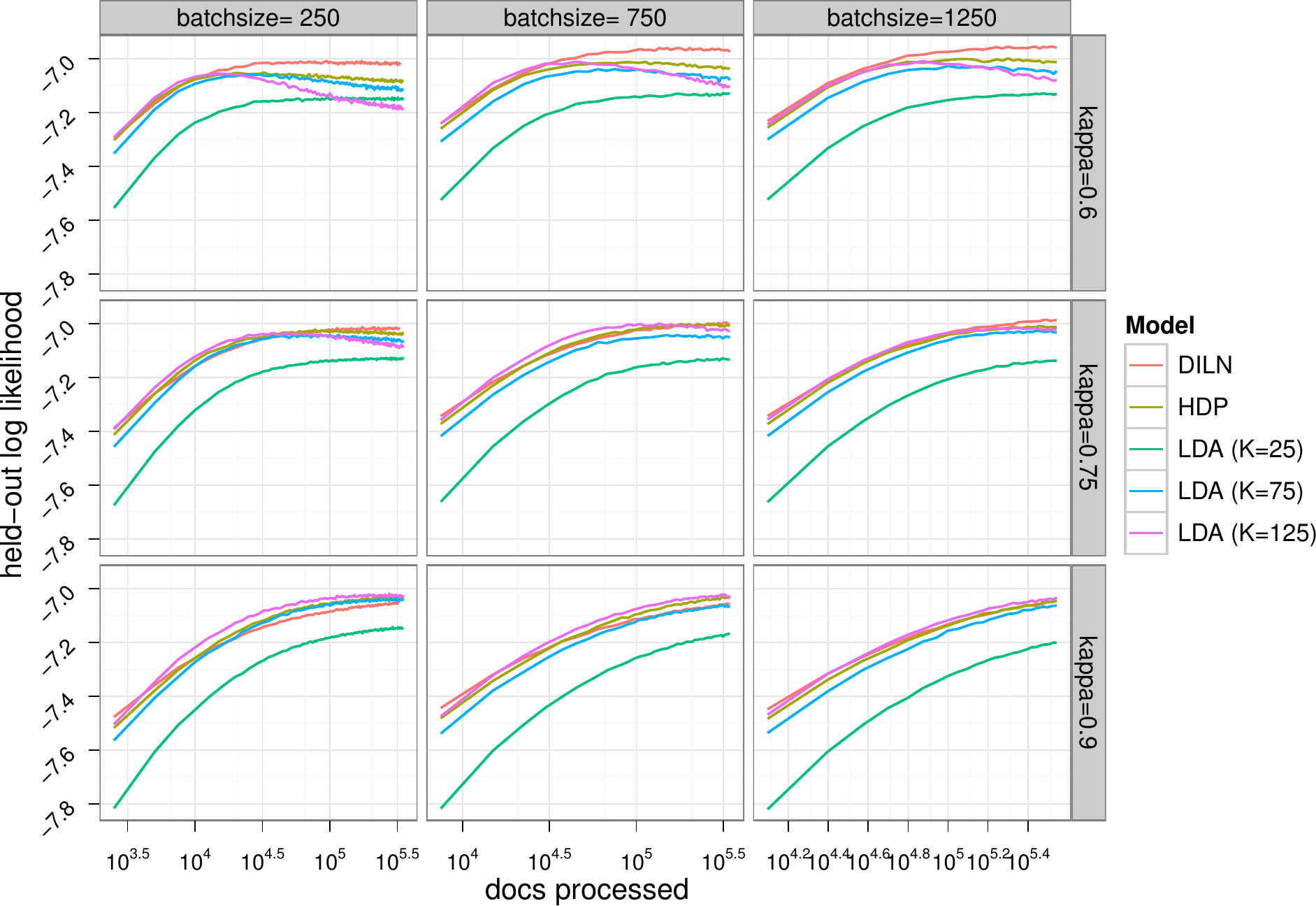}}
\caption{Stochastic variational inference results on \textit{Nature}. The number of documents processed is shown in log scale. We observe improved performance for all algorithms as $\kappa$ decreases, and note that DILN is able to obtain a level of performance not reached by HDP and LDA as a function of parameter settings.}\label{fig.onlineresults} 
\end{figure}

\paragraph{Stochastic variational inference}

We compare stochastic DILN with stochastic HDP and online LDA using 352,549 documents from \emph{Nature}. As for
batch inference, we can obtain a stochastic inference algorithm for
the HDP as a special case of stochastic DILN.  In DILN, we again use a
latent space of $d=20$ dimensions for the component locations and set the
location variance parameter to $c = 1/20$.  We truncate the models at
$200$ topics, and we evaluate performance for $K \in \{25, 75, 125\}$
topics with stochastic inference for LDA~\citep{Hoffman:2010}.  As we
discussed in Section \ref{sec.online}, we use a step sequence of
$\rho_t = (\zeta + t)^{-\kappa}$.  We set $\zeta = 25$, and run the
algorithm for $\kappa \in \{0.6, 0.75, 0.9\}$.  We explored various
batch sizes, running the algorithm for $|B_t| \in \{250, 750, 1250\}$.
Following \cite{Hoffman:2010}, we set the topic Dirichlet
hyperparameters to $\gamma_0 = 0.01$.

For testing, we held out $10,000$ randomly selected documents from the
corpus.  We measure the performance of the stochastic models after
every $10$th batch.  Within each batch, we run several iterations of
local variational inference to find document-specific parameters.  We
update corpus-level parameters when the change in the average per-document topic distributions falls below a threshold. On average, roughly ten document-level iterations were run for each corpus-level update.

\begin{figure}[t!]
\includegraphics[width=.99\textwidth]{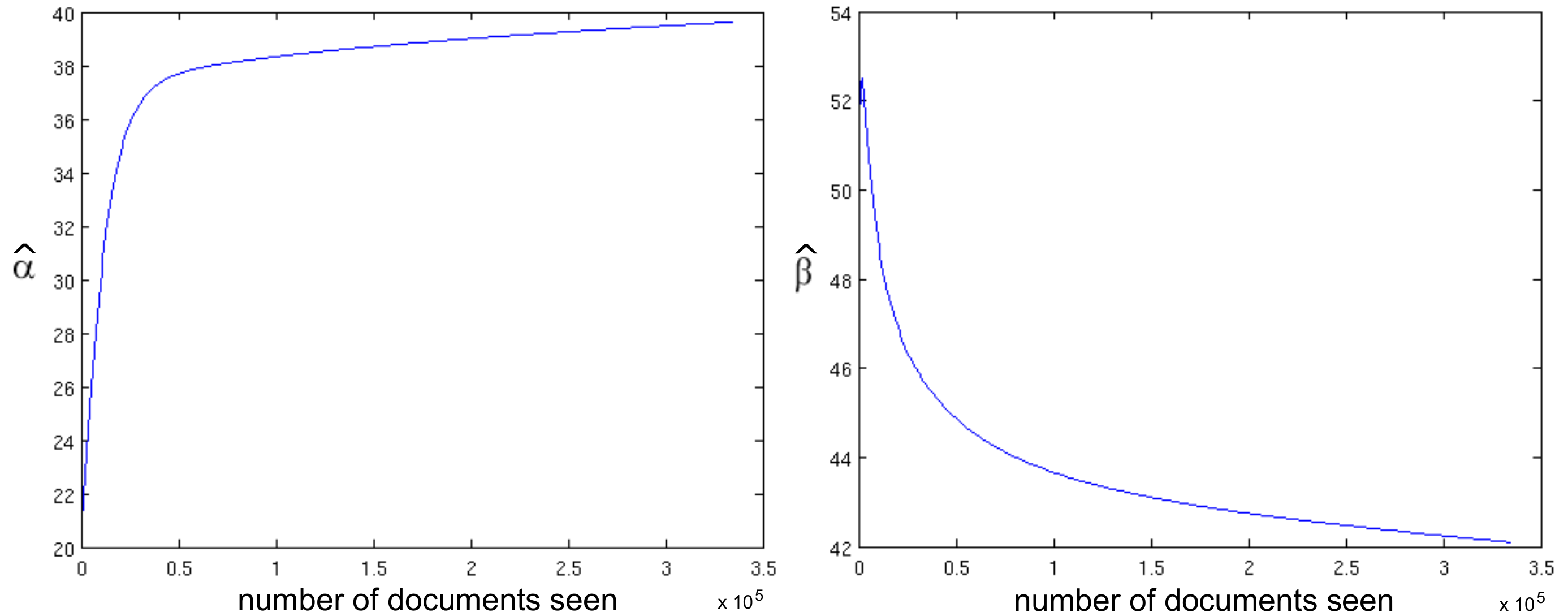}
\caption{Stochastic learning of \textit{Nature}. The values of $\hat{\alpha}$ and $\hat{\beta}$ as a function of number of documents seen for batch size equal to 750 and learning rate $\kappa = 0.6$.}\label{fig.alphabeta}\vspace{5mm}
\includegraphics[width=.99\textwidth]{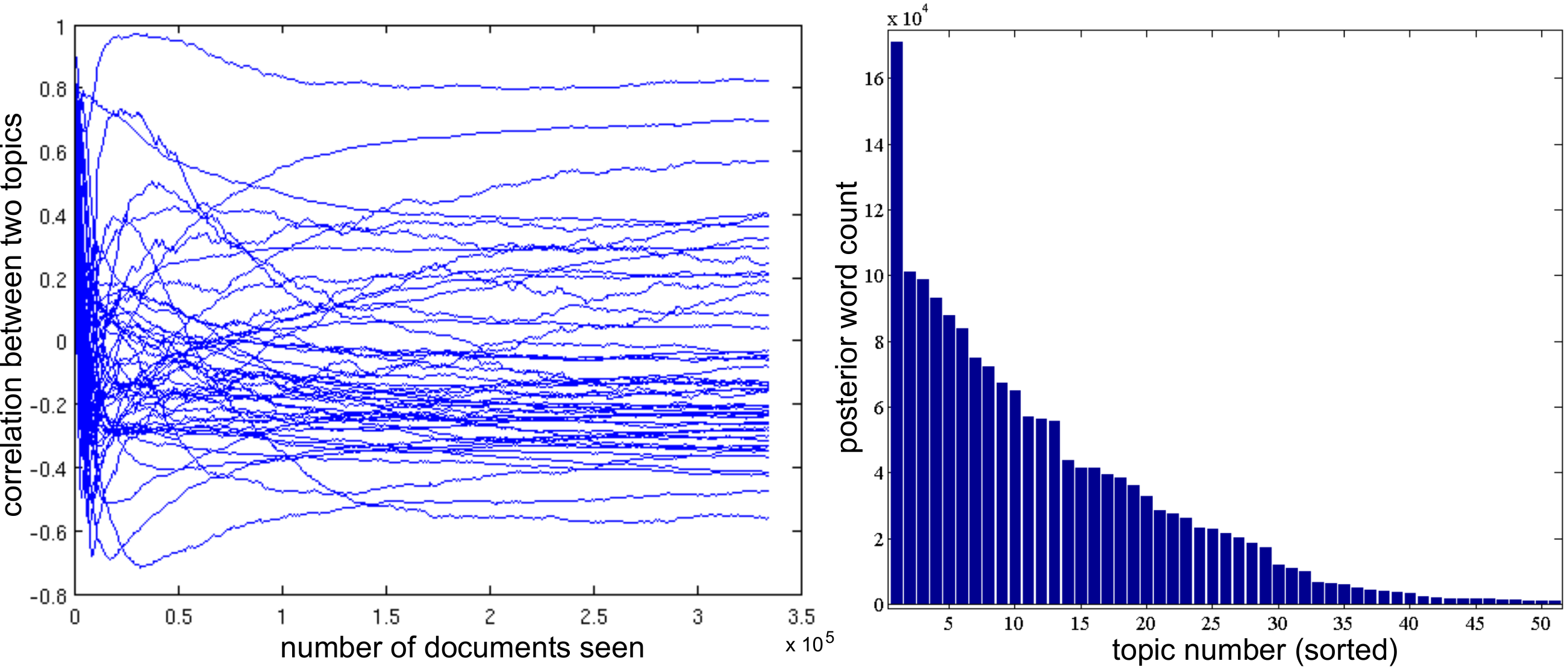}
\caption{Stochastic learning of \textit{Nature}. (left) Correlations between 100 randomly selected pairs of topics as a function of documents seen. (right) The empirical word count from the posteriors of the top 50 topics after the final iteration. Approximately 50 of the 200 topics are used.}\label{fig.otherstuff}
\end{figure}

Figure \ref{fig.onlineresults} illustrates the results.  In this
figure, we show the per-word held-out perplexity as a function of the
number of documents seen by the algorithm. From these plots we see
that a slower decay in the step size improves performance. Especially
for DILN, we see that performance improves significantly as the decay
$\kappa$ decreases, since more information is being used from later
documents in finding a maximum of the variational objective
function. Slower decays are helpful because more parameters are being
fitted by DILN than by the HDP and LDA.  We observed that as $\kappa$
increases a less detailed correlation structure was found; this
accounts for the decrease in performance.

\begin{figure}
\includegraphics[width=.95\textwidth]{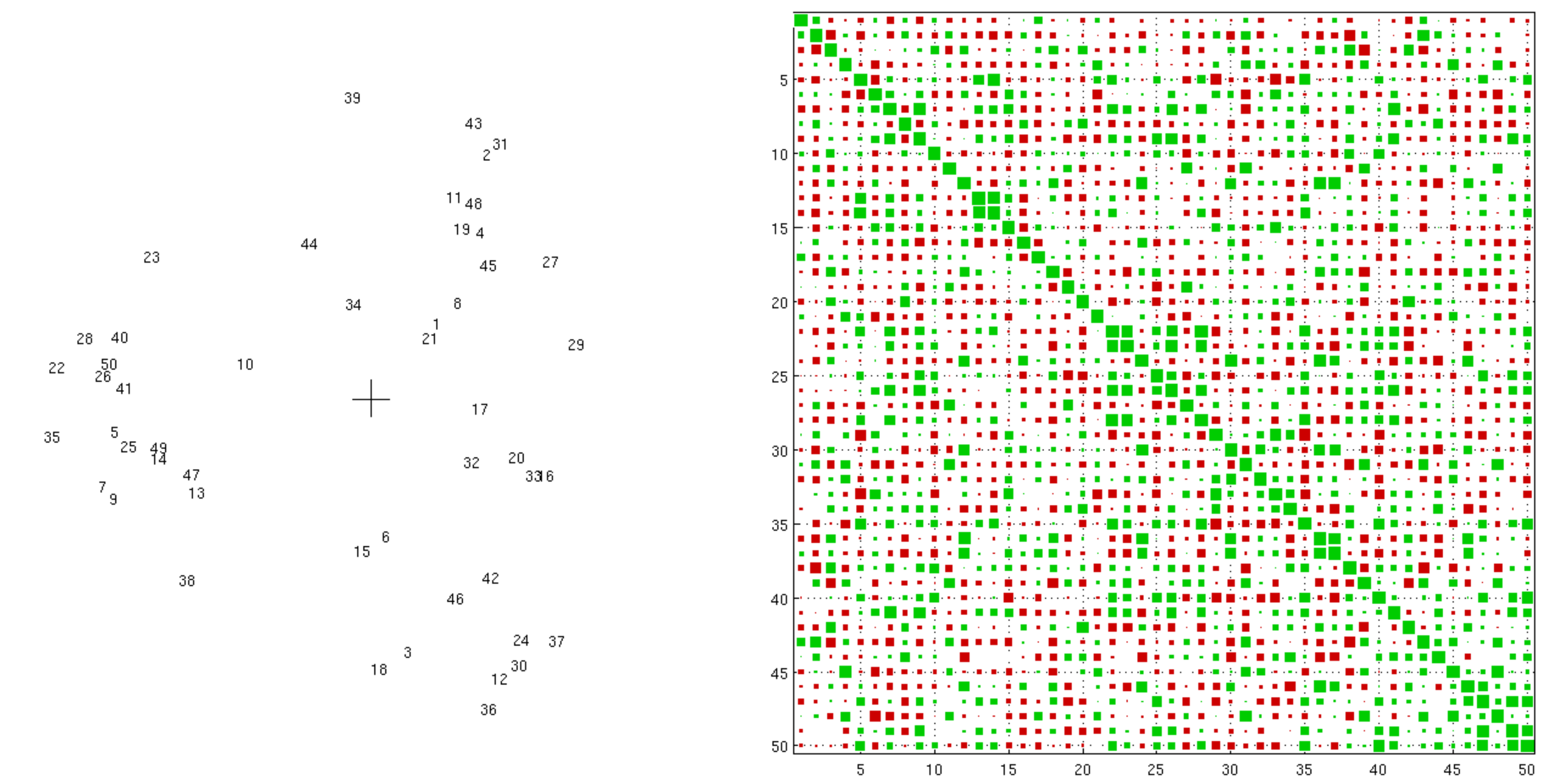}\vspace{1mm}
\begin{scriptsize}
\begin{tabular}{l}
Topic 1: author, facts, original, written, hand, text, think, himself, pages, mind\\
Topic 2: war, england, carried, death, french, german, issued, great-britain, sent, works\\
Topic 3: equation, flow, sample, average, mantle, rates, distribution, zone, ratios, calculated\\
Topic 4: million, scientists, policy, britain, social, economic, technology, political, project, organization\\
Topic 5: gene, genes, expression, mutant, wild-type, sequence, supplementary, embryos, mutants, clones\\
Topic 6: glass, tube, colour, due, substance, rays, apparatus, substances, action-of, series\\
Topic 7: serum, labelled, fraction, anti, purified, buffer, fractions, rabbit, extract, extracts\\
Topic 8: feet, rocks, island, specimens, sea, coast, islands, river, land, geological\\
Topic 9: membrane, enzyme, concentration, glucose, inhibition, calcium, release, phosphate\\ 
Topic 10: population, evolution, selection, genetic, environment, evolutionary, food, birds, breeding\\
Topic 11: college, secretary, council, cambridge, department, engineering, assistant, mathematics\\ 
Topic 12: frequency, wave, spectrum, electron, absorption, band, electrons, optical, signal, peak\\
Topic 13: binding, proteins, residues, peptide, chain, amino-acid, domain, terminal, sequence\\
Topic 14: dna, rna, sequence, sequences, mrna, poly, fragments, synthesis, fragment, phage\\
Topic 15: molecules, compounds, oxygen, molecule, reactions, formation, ion, ions, oxidation, compound\\
Topic 16: the-sun, solar, the-earth, motion, observatory, stars, comet, star, night, planet\\
Topic 17: techniques, materials, applications, reader, design, basic, service, computer, fundamental\\ 
Topic 18: crystal, structures, unit, orientation, ray, diffraction, patterns, lattice, layer, symmetry\\
Topic 19: vol, museum, plates, india, journal, ltd, net, indian, series, washington\\
Topic 20: sea, ice, ocean, depth, deep, the-earth, climate, sediments, earth, global\\
Topic 21: you, says, her, she, researchers, your, scientists, colleagues, get, biology\\
Topic 22: mice, anti, mouse, tumour, antigen, antibody, cancer, tumours, antibodies, antigens\\
Topic 23: disease, blood, bacteria, patients, drug, diseases, clinical, drugs, bacterial, host\\
Topic 24: radio, ray, emission, flux, stars, disk, sources, star, galaxies, galaxy\\
Topic 25: brain, receptor, receptors, responses, stimulation, response, stimulus, cortex, synaptic, stimuli\\
Topic 26: rats, liver, tissue, blood, dose, injection, rat, plasma, injected, hormone\\
Topic 27: royal, lecture, lectures, engineers, royal-society, hall, institution-of, society-at, annual, january\\
Topic 28: virus, cultures, culture, medium, infected, infection, viral, viruses, agar, colonies\\
Topic 29: heat, oil, coal, electric, electricity, electrical, lead, supply, steam, tons\\
Topic 30: particles, particle, electron, proton, neutron, protons, mev, force, scattering, nuclei\\
Topic 31: education, universities, training, schools, teaching, teachers, courses, colleges, grants, student\\
Topic 32: nuclear, radiation, irradiation, radioactive, uranium, fusion, reactor, storage, damage\\
Topic 33: iron, copper, steel, metals, milk, aluminium, alloys, silicon, ore, haem\\
Topic 34: soil, nitrogen, leaves, land, agricultural, agriculture, nutrient, yield, growing, content\\
Topic 35: chromosome, nuclei, hybrid, chromatin, mitotic, division, mitosis, chromosomal, somatic\\
Topic 36: pulse, spin, magnetic-field, pulses, polarization, orbital, decay, dipole, pulsar, polarized\\
Topic 37: atoms, quantum, atom, einstein, classical, photon, relativity, bohr, quantum-mechanics\\
Topic 38: strain, stress, strains, deformation, shear, stresses, failure, viscosity, mechanical, stressed\\
Topic 39: medical, health, medicine, tuberculosis, schools, education, teaching, infection, bacilli, based\\
Topic 40: adult, females, males, mating, mature, progeny, adults, maturation, aggressive, matings
\end{tabular}
\end{scriptsize}
\caption{Stochastic DILN after one pass through the \textit{Nature} corpus. The upper left figure shows the \textit{projected} topic locations with + marking the origin. The upper right figure shows topic correlations. We list the ten most probable for the first 40 topics.}\label{fig.natwords}
\end{figure}

In Figure \ref{fig.natwords} we show the model after one pass through
the \emph{Nature} corpus. The upper left figure shows the locations of
the top 50 topics projected from $\mathbb{R}^{20}$. These locations
are rough approximations since the singular values were large for
higher dimensions. The upper right figure shows the correlations
between the topics. Below these two plots, we show the ten most
probable words from the 50 most probable topics. In Figure
\ref{fig.alphabeta} we show $\hat{\alpha}$ and $\hat{\beta}$ as a
function of the number of documents seen by the model. In Figure
\ref{fig.otherstuff} we show the correlations between 100 pairs of
topics chosen at random; these are also shown as a function of the
number of documents seen. In general, these plots indicate that the
parameters are far along in the process of converging to a local
optimum after just one pass through the entire corpus. Also shown in
Figure \ref{fig.otherstuff} is the empirical word count per topic (that is, the values $\sum_{m,n} \mathbb{I}(C_n^{(m)} = k)$ as a function of $k$)
after the final iteration of the first pass through the data. We see
that the model learns approximately 50 topics out of the 200 initially
supplied. All results are shown for a batch size of 750.

\paragraph{Stochastic DILN vs batch DILN}

We also compare stochastic and batch inference for DILN to show how
stochastic inference can significantly speed up the inference process,
while still giving results as good as batch inference. We again use
the \emph{Nature} corpus. For stochastic inference, we use a subset of
size $|B_t| = 1000$ and a step of $(1+t)^{-0.75}$. For batch
inference, we use a randomly selected subset of documents, performing
experiments on corpus size $M \in \{25000,50000,100000\}$. All
algorithms used the same test set and testing procedure, as discussed
in Section \ref{sec.evaluation}. All experiments were run on the same
computer to allow for fair time comparisons.

\begin{figure}
 \includegraphics[width=1\textwidth]{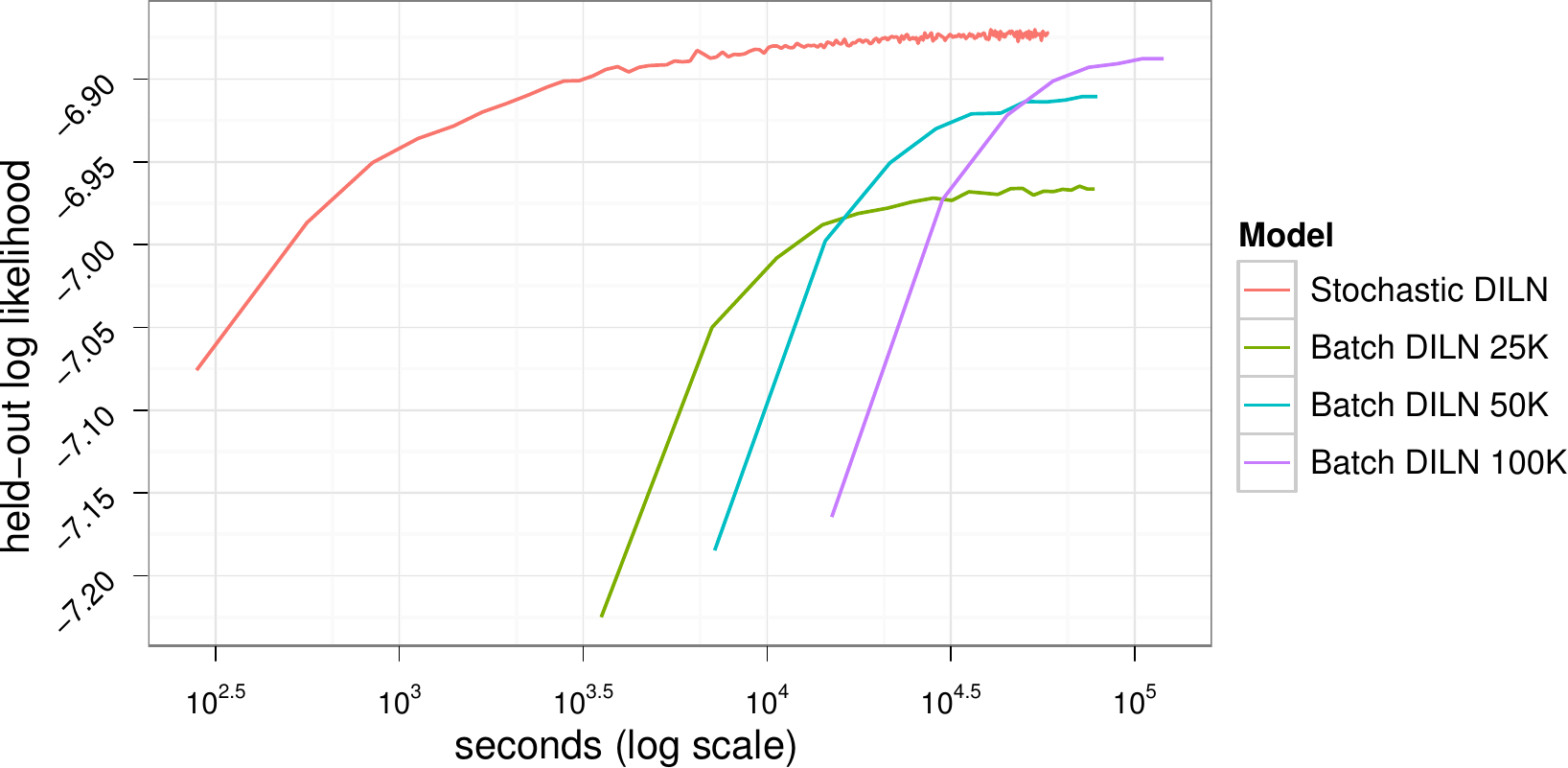}
\caption{A comparison of stochastic and batch inference for DILN using the \emph{Nature} corpus. Results are shown as a function of time (log scale). Stochastic inference achieves a good posterior approximation significantly faster than batch inference, which pays for improved performance with an increasing runtime.}\label{fig.onlinevsbatch}
\end{figure}

In Figure \ref{fig.onlinevsbatch}, we plot the held-out per-word log
likelihood as a function of time. We measured performance every tenth iteration to construct each curve. The stochastic inference curve
represents roughly six passes through the entire corpus. For batch inference, we see that performance improves significantly as the sub-sampled batch size increases. However, this improvement is paid for with an increasing runtime. Stochastic inference is much faster, but still performs as well as batch in predicting test documents.

\section{Discussion}

We have presented the discrete infinite logistic normal distribution,
a Bayesian nonparametric prior for mixed-membership models. DILN
overcomes the hidden assumptions of the HDP and explicitly models
correlation structure between the mixing weights at the group
level. We showed how using the second parameter of the gamma process
representation of the hierarchical Dirichlet process achieves this by
varying per-component according to an exponentiated Gaussian
process. This Gaussian process is defined on latent component
locations added to the hierarchical structure of the HDP.

Using batch variational Bayesian inference, we showed an improvement
in predictive ability over the HDP and the CTM in a topic modeling
application. Furthermore, we showed how this algorithm can be modified
to obtain a new variational inference algorithm for HDPs based on the
gamma process. We then extended the model to the stochastic inference
setting, which allows for fast analysis of much larger corpora.

DILN can be useful in other modeling frameworks. For example, hidden
Markov models can be viewed as a collection of mixture models that are
defined over a shared set of parameters, where state transitions
follow a Markov transition rule. \cite{Teh:2006} showed how the HDP
can be applied to the HMM to allow for infinite state support, thus
creating a nonparametric hidden Markov model, where the number of
underlying states is inferred. DILN can be adapted to this problem as
well, in this case modeling correlations between state transition
probabilities.

\section{Appendix}
\subsection{Proof of almost sure finiteness of $\sum_{i=1}^{\infty} Z_i\e^{w_i}$}
We drop the group index $m$ and define $w_i := W(\ell_i)$. The normalizing constant for DILN, prior to absorbing the scaling factor within the gamma distribution, is $S := \sum_{i=1}^{\infty} Z_i\e^{w_i}$. We first show that this value is finite almost surely when the Gaussian process has bounded mean and covariance functions. This case would apply for example when using a Gaussian kernel. We then give a proof for the kernel in Section \ref{sec:kernel} when the value of $c < 1$.

Let $S_T := \sum_{i=1}^T Z_i\e^{w_i}$. It follows that $S_1 \leq \dots \leq S_T \leq \dots \leq S$ and $S = \lim_{T \to\infty}S_T$. To prove that $S$ is finite almost surely, we only need to prove that $\mathbb{E}[S]$ is finite. From the monotone convergence theorem, we have that $\mathbb{E}[S] = \lim_{T\to \infty}\mathbb{E}[S_T]$. Furthermore, $\mathbb{E}[S_T]$ can be upper bounded as follows,
\begin{equation}
\mathbb{E}[S_T] = \textstyle \sum_{i=1}^T \mathbb{E}[Z_i] \mathbb{E}[\e^{w_i}]\leq \e^{\max_i (\mu_i + \frac{1}{2}\sigma_i^2)}\textstyle\sum_{i=1}^T \mathbb{E}[Z_i].
\end{equation}
$\mathbb{E}[S]$ is therefore upper bounded by $\beta \e^{\max_i (\mu_i + \frac{1}{2}\sigma_i^2)}$ and $S$ is finite almost surely.

For the kernel in Section \ref{sec:kernel}, we prove that $\mathbb{E}[S] < \infty$ when $c < 1$. We only focus on this case since values of $c \geq 1$ are larger than we are interested in for our application. For example, given that $\ell \in \mathbb{R}^d$ and $\ell \sim \mathrm{Normal}(0,cI_d)$, it follows that $\mathbb{E}[\ell^T\ell] = dc$, which is the expected variance of the Gaussian process at this location. In our applications, we set $c = 1/d$, which is less than one when $d>1$. As above, we have
\begin{equation}
 \mathbb{E}[S_T] = \textstyle \sum_{i=1}^T \mathbb{E}[Z_i] \mathbb{E}[\e^{\ell_i^T u}] = \textstyle \sum_{i=1}^T \beta p_i \mathbb{E}[\e^{\frac{c}{2}u^T u}].
\end{equation}
Since $u\sim\mathrm{Normal}(0,I_d)$, this last expectation is finite when $c < 1$, and therefore the limit $\lim_{T\rightarrow\infty} \mathbb{E}[S_T]$ is also finite.

\subsection{Variational inference for normalized gamma measures}\label{sec:vbapprox}
In DILN, and normalized gamma models in general, the expectation of the log of the normalizing constant, $\bbE_Q[\ln \sum_k Z_k]$, is intractable. We present a method for approximate variational Bayesian inference for these models. A Taylor expansion on this term about a particular point allows for tractable expecations, while still preserving the lower bound on the log-evidence of the model. Since the log function is concave, the negative of this function can be lower bounded by a first-order Taylor expansion,
\begin{equation}\label{eqn.taylor1}
-\bbE_Q\left[\ln \sum_{k=1}^T Z_k\right] \geq -\ln \xi - \frac{\sum_k\bbE_Q[Z_k] - \xi}{\xi}.
\end{equation}
We have dropped the group index $m$ for clarity. A new term $\xi$ is introduced into the model as an auxiliary parameter. Changing this parameter changes the tightness of the lower bound, and in fact, it can be removed by permanently tightening it,
\begin{equation}\label{eqn.lbparam}
\xi = \sum_{k=1}^T\bbE_Q[Z_k].
\end{equation}
In this case $\bbE_Q[\ln \sum_k Z_k]$ is replaced with $\ln \sum_k \bbE_Q [Z_k]$ in the variational objective function. We do not do this, however, since retaining $\xi$ in DILN allows for analytical parameter updates, while using Equation (\ref{eqn.lbparam}) requires gradient methods. These analytical updates result in an algorithm that is significantly faster. For example, inference for the corpora considered in this paper ran approximately five times faster. 

Because this property extends to variational inference for all mixture models using the normalized gamma construction, most notably the HDP, we derive these updates using a generic parameterization of the gamma distribution, $\mathrm{Gamma}(a_k,b_k)$. The posterior of $Z_{1:T}$ in this model is proportional to
\begin{equation}
p(Z_{1:T}|C_{1:N},a_{1:T},b_{1:T}) \propto \left[\prod_{n=1}^N \prod_{k=1}^T \left(\frac{Z_k}{\sum_j Z_j}\right)^{\I(C_n=k)}\right]\left[
\prod_{k=1}^T Z_k^{a_k-1}\e^{-b_k Z_k}\right].
\end{equation}
Under a factorized $Q$ distribution, the variational lower bound at nodes $Z_{1:T}$ is
\begin{eqnarray}\label{eqn.VBLB1}
\bbE_Q[\ln p(Z_{1:T}|-)] + \bbH[Q] &=& \sum_{n=1}^N\sum_{k=1}^T \bbP_Q(C_n = k)\bbE_Q[\ln Z_k] - N\bbE_Q\left[\ln \sum_{k=1}^T Z_k\right]\nn\\
&& + \sum_{k=1}^T  (\bbE_Q[a_k]-1)\bbE_Q[\ln Z_k] -  \sum_{k=1}^T \bbE_Q[b_k]\bbE_Q[Z_k] \nn\\
&& + \sum_{k=1}^T\bbH[Q(Z_k)] \,+\, \mbox{const.}
\end{eqnarray}
The intractable term, $- N\bbE_Q[\ln \sum_k Z_k]$, is replaced with the bound in Equation (\ref{eqn.taylor1}).

Rather than calculate for a specific $q$ distribution on $Z_k$, we use the procedure discussed by \cite{Winn:2005} for finding the optimal form and parameterization of a given $q$: We exponentiate the variational lower bound in Equation (\ref{eqn.VBLB1}) with all expectations involving the parameter of interest not taken. For $Z_k$, this gives
\begin{eqnarray}
q(Z_k) &\propto& \e^{\bbE_{Q_{-Z_k}}[\ln p(Z_k|C_{1:N},a_{1:T},b_{1:T})]}\nn\\
       &\propto& Z_k^{\bbE_Q[a_k] + \sum_{n=1}^N \bbP_Q(C_n = k)-1}~\e^{-(\bbE_Q[b_k] + N/\xi)Z_k}.
\end{eqnarray}
Therefore, the optimal $q$ distribution for $Z_k$ is $q(Z_k) = \mathrm{Gamma}(Z_k|a'_k, b'_k)$ with $a'_k = \bbE_Q[a_k] + \sum_{n=1}^N \bbP_Q(C_n = k)$ and $b'_k = \bbE_Q[b_k] + N/\xi$. The specific values of $a'_k$ and $b'_k$ for DILN are given in Equation (\ref{eqn.upZ}).

\bibliographystyle{imsart-nameyear}
\bibliography{DILN}

\end{document}